\documentclass{article}
\usepackage{iclr2027_conference,times}

\usepackage{amsmath,amsfonts,bm}

\def\eqref#1{equation~\ref{#1}}

\def\1{\bm{1}}

\DeclareMathAlphabet{\mathsfit}{\encodingdefault}{\sfdefault}{m}{sl}
\SetMathAlphabet{\mathsfit}{bold}{\encodingdefault}{\sfdefault}{bx}{n}

\usepackage{hyperref}
\usepackage{url}
\usepackage{booktabs}
\usepackage{multirow}
\usepackage{tabularx}
\usepackage{xspace}
\usepackage{xcolor}
\usepackage{graphicx}
\usepackage{wrapfig}
\usepackage{needspace}
\usepackage{caption}
\usepackage{subcaption}\usepackage{siunitx}
\usepackage{enumitem}
\usepackage{amsmath}
\usepackage{amsthm}

\theoremstyle{definition}

\newcommand{\pp}{\,\textrm{pp}}

\title{When Upstream Messages Override Correct Answers: A Controlled Study of Multi-Agent LLM Collaboration}

\author{Yaxin Gong$^{1,2}$, Gangyi Zhang$^{2}$, Chongming Gao$^{1}$, Leyang Shen$^{3}$, Chenxiao Fan$^{1}$,
\\
\textbf{Jiakai Wang$^{2}$, Dong Wang$^{2}$, Yang Liu$^{2}$, Wenjie Wang$^{1}$, Xiangnan He$^{1}$}\\
$^{1}$ University of Science and Technology of China \\ 
$^{2}$ Qwen Business Unit of Alibaba \quad $^{3}$ National University of Singapore \\ 
\texttt{gyx2022@mail.ustc.edu.cn} \\
}

\iclrfinalcopy

\begin{document}
\maketitle
\lhead{Preprint}

\begin{abstract}
Multi-agent LLM systems rely on message passing among specialized agents to accomplish complex tasks. However, an upstream agent may provide useful information or an incorrect answer that causes a downstream agent to override a correct answer supported by its own evidence. Prior work has not clearly separated the benefits of communication from the damage caused by incorrect messages.
We study this problem with controlled experiments across five benchmarks and five receivers, keeping the downstream task and evidence fixed while comparing answers under three conditions: no message, the upstream agent's original message, or a message with the opposite conclusion.
Our experiments reveal three key findings. First, messages often help when the downstream agent would otherwise answer incorrectly. Second, messages can also hurt: when the downstream agent would answer correctly without a message, an incorrect upstream message changes the answer in up to 32\% of cases. Third, in 94\% of audited harmful cases, the downstream agent copies the upstream's specific wrong answer---a pattern we term \emph{answer substitution}.
Removing unreliable messages recovers part of the lost accuracy, suggesting that communication should be selective based on upstream reliability and the evidence already available to the downstream agent.
\end{abstract}

\section{Introduction}
\label{sec:intro}

Multi-agent systems distribute complex tasks among specialized agents that collaborate through message passing, and have been widely adopted in code generation~\citep{ICLR2024_6507b115,wu2024autogen}, question answering~\citep{zhang2024aflow,hu2024adas,chen2024agentverse}, reasoning and debate~\citep{du2024debate,liang2024rethinking,chan2024chateval}, and tool-augmented workflows~\citep{NEURIPS2023_d842425e,yao2023react}.

\begin{figure*}[t]
  \centering
  \includegraphics[width=\textwidth]{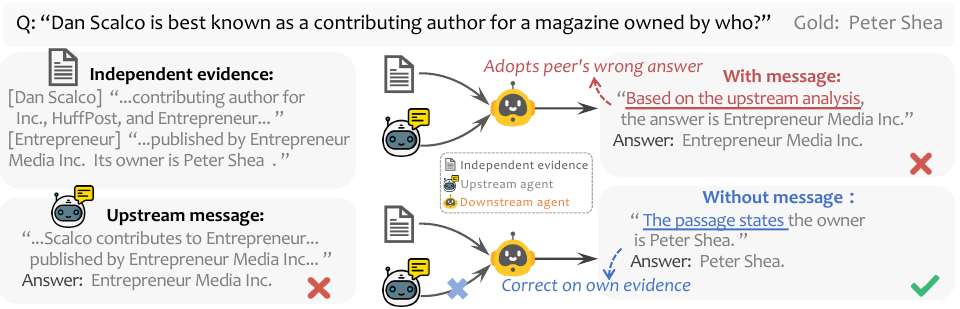}
  \caption{An erroneous peer message overrides an otherwise correct, evidence-supported answer.}
  \label{fig:intro}
\end{figure*}

However, errors and noise are pervasive in multi-agent communication~\citep{cemri2025mast,zhang2026hallucination,pan2026misinformation}.
Upstream agents produce hallucinations~\citep{zhang2026hallucination,guan2026snowball}, reasoning mistakes~\citep{cemri2025mast}, or formatting errors, and these errors propagate downstream through messages~\citep{pan2026misinformation,yang2026wrongbutuseful}.
More critically, even when the downstream agent holds sufficient independent evidence, an erroneous upstream message can still override its otherwise correct judgment~\citep{qu2026conformity,cho2025herd,wan2026deliberative}. Figure~\ref{fig:intro} illustrates a concrete case: the receiver cites the correct evidence yet adopts the peer's wrong answer.

Prior work has approached this problem from several angles.
Conformity studies show that simulated peer opinions can mislead models~\citep{qu2026conformity,cho2025herd}; sycophancy research finds that aligned models change correct answers to match user preferences~\citep{sharma2024sycophancy,wei2023sycophancy}; and knowledge-conflict studies examine contradictions between what a model has learned and what the context provides~\citep{xie2024knowledge,chen2022rich}.
However, these studies are largely based on single-model observations or simulated voting scenarios, lacking controlled causal analysis that fixes evidence and manipulates messages in real pipelines.
In this paper, we systematically analyze the reliability of multi-agent communication through controlled experiments.
We let the downstream agent answer QA and code-generation tasks independently, then compare how its answers change when the upstream message is hidden, shown as-is, or conclusion-reversed---measuring item by item how each message affects downstream judgment.
We cross five benchmarks with five receivers, and rule out alternative explanations through matched controls that vary the message's source label, reception timing, and scoring method.

We find that peer messages are generally helpful: when the downstream agent would otherwise answer incorrectly, the upstream message helps it.
But when the upstream is wrong, that message causes the downstream agent to abandon its correct answer---up to 32\% of correct answers are overridden when the message is present (Figure~\ref{fig:intro}). In 94\% of audited cases, the override targets the upstream's specific wrong answer.
We call this pattern \emph{substitution}: the downstream agent holds sufficient evidence to answer correctly, yet still follows the upstream's error.
The matched controls described above confirm that substitution is driven by the message content itself, not by the message's position, source label, or the additional reasoning opportunity it provides. 
The effect is modulated by conditions: when the receiver holds independent evidence, the marginal value of the message is close to zero (help and harm nearly cancel); more accurate upstream models reduce the number of affected items but do not eliminate per-item displacement.

Our analysis also points toward mitigation: when the system can detect that the upstream message is likely wrong, removing the message or replacing the downstream model can recover a portion of the overridden correct answers, suggesting that communication protocol design is a key lever for improving multi-agent reliability.

Our contributions:
\begin{enumerate}[nosep,leftmargin=*]
  \item We identify an underexplored reliability problem in multi-agent communication: even when the downstream agent holds sufficient evidence, a single erroneous upstream message can cause it to abandon its correct answer. Prior studies of conformity and knowledge conflict have not isolated this effect with item-level designs that fix evidence and manipulate message content in real pipelines.
  \item We design controlled experiments that fix downstream evidence and vary only the upstream message across five benchmarks and five receivers, ruling out multiple alternative explanations to characterize the problem along three dimensions: overall message value, per-item displacement direction, and boundary conditions.
  \item We find that up to 32\% of correct answers are overridden when the message is present, with 94\% of audited cases targeting the upstream's specific wrong answer. Detecting and removing erroneous messages can partially recover these losses, indicating that selective message gating is a viable lever for improving pipeline reliability.
\end{enumerate}

\section{Analysis}
\label{sec:setup}

\subsection{Experimental Design}
\label{sec:setup-framing}

We study a single \emph{message reception} in a draft-review handoff: an upstream agent produces a message~$m$ on a subtask, then a downstream agent receives~$m$ together with task-relevant \emph{evidence}~$e$ (retrieved passages, database schemas, or tool outputs~\citep{NEURIPS2023_d842425e}) and produces a final answer.
This two-node handoff is a common building block of sequential multi-agent pipelines---automated architecture search via AFlow~\citep{zhang2024aflow} converges to a two-node draft-review topology on multiple benchmarks, while others favor majority-vote aggregation (Appendix~\ref{sec:app-real-pipeline}). Our controlled design lets us attribute downstream behavior to the message content rather than to pipeline-level confounds.

\paragraph{Two measurable quantities.}
We vary two factors, whether the receiver holds independent evidence and whether it sees the peer message, forming a $2 \times 2$ factorial design.
\textbf{Message value} under evidence condition $s$ is $\tau(s) = \operatorname{Acc}_{\rm shown}(s) - \operatorname{Acc}_{\rm hidden}(s)$:
positive $\tau$ means the message helps; negative $\tau$ means it hurts.
The \textbf{evidence--message interaction} is
$\Gamma = \tau(\text{no evidence}) - \tau(\text{with evidence})$.
If $\Gamma > 0$, giving the receiver its own evidence makes the peer message less useful, or actively harmful.

\paragraph{Operational definition of substitution.}
We define substitution as an observable behavioral pattern: when the upstream message contains an incorrect answer, the receiver would have answered correctly without that message, but after receiving it, switches to the peer's specific wrong answer.
The experiments below test whether this pattern occurs and under what conditions.
We classify an item as independently solvable only if the receiver answers correctly in at least 2 of 3 independent evidence-only runs ($k{=}3$ majority vote; robustness comparison in Appendix~\ref{sec:app-stability}).

\paragraph{Theoretical motivation.}
Value-of-information theory~\citep{blackwell1953equivalent,howard1966voi} states that an additional signal cannot hurt a decision-maker who can freely ignore it.
This provides a baseline expectation for our diagnosis: if the receiver can freely dispose of the peer message, $\tau$ should be non-negative.
$\tau < 0$ means the free-disposal condition fails in practice: the receiver cannot effectively ignore an erroneous message.
$\Gamma$ measures the extent to which independent evidence changes this picture.
We retain the VoI framing as an intuitive reference point; the contribution of this paper is empirical, not theoretical.

\subsection{Research Questions}
\label{sec:setup-predictions}

The design above leads to three research questions:

\begin{enumerate}[nosep,leftmargin=*]
  \item When the upstream errs, does the message turn from helpful to harmful? We first validate that evidence reduces message value ($\Gamma > 0$, as a sanity check on the experimental design), then decompose $\tau$ by upstream correctness to reveal how the same message has opposite effects under different conditions. (Tested in \S\ref{sec:results-where}.)
  \item Is the displacement directional? On items the receiver can answer correctly alone, does the message cause it to switch to the peer's specific wrong answer, not merely a random error? (Tested in \S\ref{sec:results-harm}.)
  \item Does the answer follow message content? Holding evidence constant and reversing the peer conclusion, does the receiver's answer shift accordingly, tracking the message rather than the evidence? (Tested in \S\ref{sec:results-flip}.)
\end{enumerate}

A positive $\Gamma$ alone does not distinguish substitution from ordinary information redundancy: if the evidence already supplies the answer, any additional signal naturally becomes less useful.
The key evidence comes from questions~2 and~3: redundancy predicts neither directional displacement nor conclusion-sensitive answer shifts.
A within-benchmark masked-evidence experiment confirms this link (\S\ref{sec:results-where}).

\subsection{Alternative Explanations to Narrow}
\label{sec:setup-alternatives}

Even if all three research questions are answered affirmatively, the behavioral pattern could in principle arise from mechanisms other than over-weighting the peer's message.
The observed displacement could also arise if the message anchors the answer simply by appearing first; the receiver defers because the message is labeled as a teammate's output; any second input triggers re-examination that randomly changes answers; or the message alters output formatting in ways that inflate apparent harm under automatic scoring.
Section~\ref{sec:results-rule-out} tests each of these alternatives with matched controls.

\section{Experiments}
\label{sec:results}

\subsection{Setup}
\label{sec:exp-setup}

We instantiate the draft-review handoff on five benchmarks: BIRD~\citep{li2024bird} (SQL generation, $n{=}150$), HotpotQA~\citep{yang2018hotpotqa} and LBMusique~\citep{trivedi2022musique} (multi-hop QA, $n{=}200$ and $160$), 2WikiMultihopQA~\citep{ho2020l2w} (multi-hop QA, $n{=}120$), and DROP~\citep{dua2019drop} (reading comprehension, $n{=}120$).
Full benchmark details are in Appendix Table~\ref{tab:hosts}.

The base design crosses two binary factors: \emph{independent evidence} (present or absent) and \emph{peer message} (shown or hidden via a neutral placeholder), yielding four conditions per item.
Additional manipulations are run as separate controlled experiments on item subsets.

The primary upstream is gpt-4o-mini; we replicate with gpt-5.4, kimi-k2.6, and qwen-plus as alternative upstreams.
The primary downstream receiver is gpt-4o-mini; cross-family generality is tested with deepseek-v3.2, kimi-k2.6, glm-5, and qwen3.6-plus (five receivers total).
All runs use temperature~0; full model and prompt details are in Appendix~\ref{sec:app-model-manifest}.

L2W has few natural upstream errors (Appendix Table~\ref{tab:hosts}); where statistical power is limited, we note this alongside results.
The interaction also replicates on a harder BIRD subset (Appendix~\ref{sec:app-bird-hard}).

\begin{figure*}[t]
  \centering
  \includegraphics[width=\textwidth]{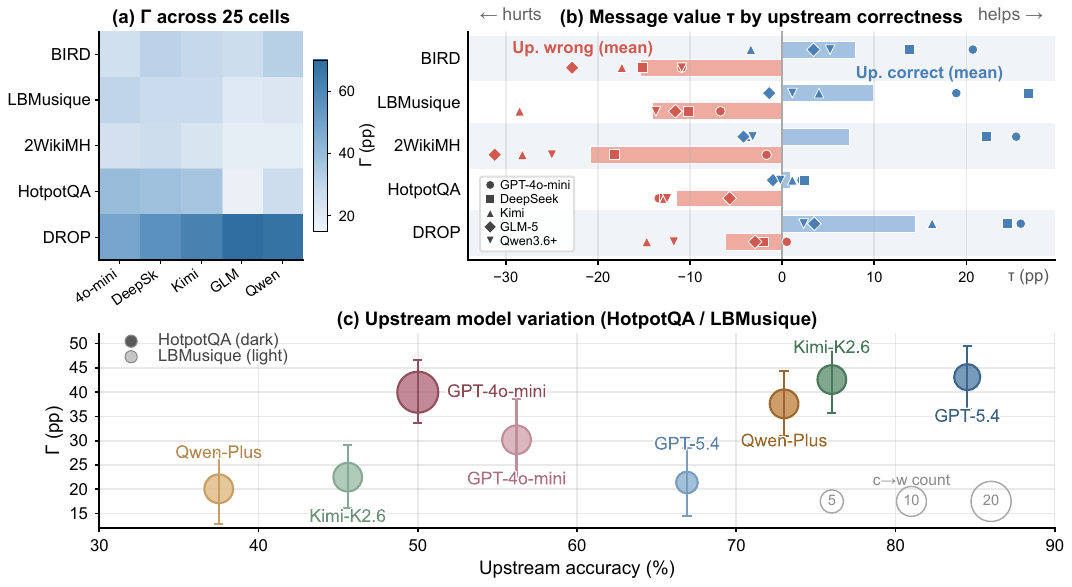}
  \caption{Evidence--message interaction, conditional message value, and upstream model variation.}
  \label{fig:q1-panel}
\end{figure*}

\subsection{The Same Message Helps and Hurts Under Different Conditions}
\label{sec:results-where}
We first test Research Question~1: when the upstream errs, does the message turn from helpful to harmful?
As a sanity check, we begin by confirming that evidence reduces message value ($\Gamma > 0$), then turn to the core finding: how $\tau$ splits by upstream correctness.

\paragraph{Sanity check: $\Gamma$ is universally positive.}
Across all 25 cells (5 receivers $\times$ 5 benchmarks), $\Gamma$ is significantly positive ($p < 0.0001$; Figure~\ref{fig:q1-panel}a).
This result is not itself surprising---when the receiver lacks evidence it can barely answer at all, so the message is obviously more valuable---but it validates the basic premise of the experimental design: evidence genuinely changes how much the receiver relies on the message, making the conditional decomposition below meaningful.

To further confirm that this interaction reflects the role of evidence itself, we mask independent evidence on BIRD (removing the database schema), observing an accuracy drop of $\Delta = -24.4$\pp (95\% CI $[-30.1, -18.8]$), ruling out correlated covariates as the source of $\Gamma$.

\paragraph{The same message both helps and hurts.}
Why is $\Gamma$ so large?
A conditional decomposition of $\tau$ by upstream correctness reveals the root cause.
Figure~\ref{fig:q1-panel}(b) shows the consistent bifurcation across all 25 cells: when the upstream answer is correct, the message helps the receiver (blue, $\tau > 0$); when the upstream answer is wrong, the same pipeline hurts the receiver (red, $\tau < 0$).
This pattern spans all five benchmarks and all five tested receivers.

\paragraph{Upstream model variation does not eliminate per-item displacement.}
Stronger upstream models make fewer errors, reducing the number of affected items.
But they do not eliminate per-item displacement.
Figure~\ref{fig:q1-panel}(c) shows results with four upstream models (accuracy 37--85\%) on HotpotQA and LBMusique: $\Gamma$ remains at 20--43\pp (all $p < 0.0001$), while the number of overridden correct answers (bubble size) shrinks with upstream capability.
Better upstreams reduce the \emph{scale} of the problem, not its \emph{nature}.

\subsection{Message Causes Directed Harm}
\label{sec:results-harm}

Research Question~1 established that the same message has opposite effects under different conditions: helpful when the upstream is correct, harmful when it errs.
We now zoom in to the micro level: is the harm directional---does it point toward the peer's specific wrong answer?

\begin{figure*}[t]
  \centering
  \includegraphics[width=\textwidth]{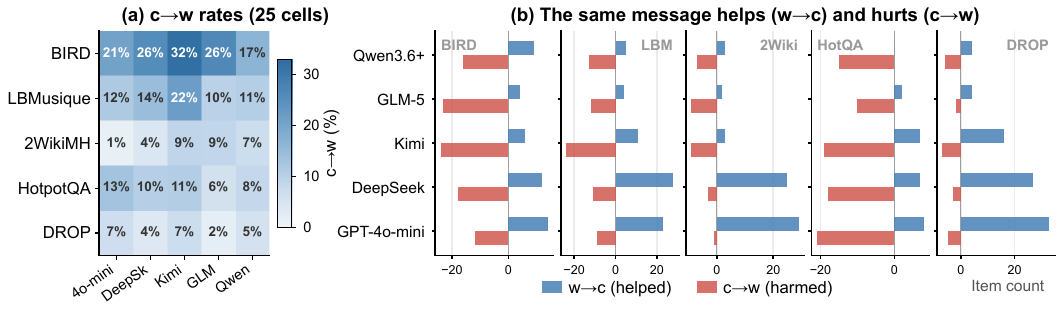}
  \caption{Override rates across 25 cells and per-benchmark transition breakdown.}
  \label{fig:q2-panel}
\end{figure*}

\paragraph{Correct answers are systematically overridden.}
Across all 25 cells, both transition types coexist: the message helps on some items (w$\to$c) while overriding correct answers on others (c$\to$w).
Figure~\ref{fig:q2-panel}(a) shows c$\to$w rates per cell: 297 of 2{,}667 independently solvable items are overridden (11.1\%), with the highest single-cell rate at 32\%.\footnote{Population counts vary by analysis scope: 297/2{,}667 covers all 25 cells (5 receivers $\times$ 5 benchmarks including DROP); the adoption audit (\S\ref{sec:results-harm}) uses 279/21 cells where per-item upstream correctness is available; the four non-DROP benchmarks total 274/2{,}184. All counts are reconciled in Appendix Table~\ref{tab:hosts}.}
Figure~\ref{fig:q2-panel}(b) shows the complementary view: the same message helps on items the receiver would otherwise fail (w$\to$c, blue) while overriding correct answers on items it could solve (c$\to$w, red).
The key finding is not that one transition type dominates, but that c$\to$w transitions are \emph{directed}---they point toward the peer's specific wrong answer.

\paragraph{Displacement is directed toward the peer's specific wrong answer.}
To confirm that the override is not random degradation, we audit 82 c$\to$w transitions.
In 77 of them (94\%), the receiver's final answer matches the specific wrong answer in the peer's message.
This is not diffuse degradation toward arbitrary errors but directed displacement toward the peer's answer---the defining feature of substitution.
Moreover, 250 of the 279 c$\to$w transitions across 21~cells (90\%) occur when the upstream answer is wrong, confirming that the displacement concentrates where the substitution definition predicts.

\paragraph{Behavioral classification.}
We classify every item with an incorrect upstream answer and independent evidence into behavioral patterns (Appendix~\ref{sec:app-taxonomy}).
The dominant pattern is \emph{unconditional following}: the receiver copies the upstream error and cannot solve the item independently either.
The most informative pattern is \emph{capable but conforming}: the receiver solves the item alone yet copies the upstream error when the message is present.
Cross-referencing with delayed receipt confirms that the vast majority of capable-but-conforming items are overwritten upon seeing the peer's wrong answer.

\subsection{The Answer Follows the Message Content}
\label{sec:results-flip}


\begin{figure*}[t]
  \centering
  \includegraphics[width=\textwidth]{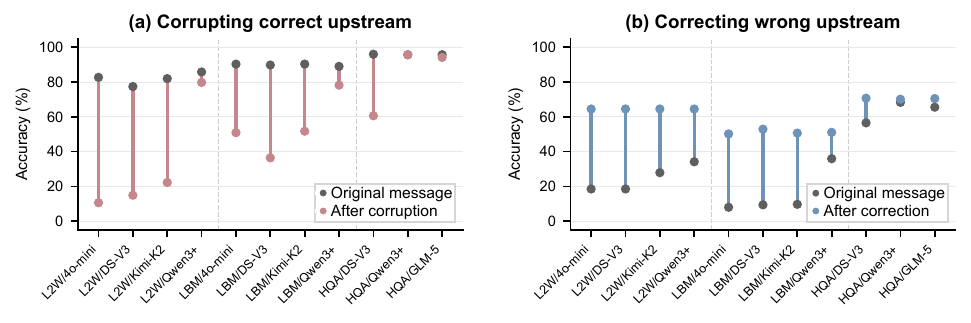}
  \caption{Cross-model conclusion reversal. \textbf{(a)}~Corrupting a correct upstream conclusion universally lowers accuracy (11/11 cells). \textbf{(b)}~Correcting a wrong conclusion universally restores it (11/11 cells). Gray dots = original message; colored dots = after reversal.}
  \label{fig:q3-reversal-main}
\end{figure*}
Research Questions~1 and~2 showed that the message harms and that the harm is directed.
Research Question~3 tests a stronger causal claim: is it the specific conclusion in the message that drives the displacement, or does the message merely cause generic interference?

\paragraph{Conclusion reversal.}
We fix the receiver's independent evidence and reverse the upstream message's conclusion---correcting an originally wrong conclusion or corrupting an originally correct one---while preserving the evidence citations and step-by-step format (60--73\% token overlap; Appendix~\ref{sec:app-flip-validation}).
Across all 11 tested cells (3--4~receivers $\times$ 3~QA benchmarks), corrupting a correct conclusion universally lowers accuracy, and correcting a wrong conclusion restores it (Figure~\ref{fig:q3-reversal-main}).
This provides strong evidence that the receiver tracks the specific conclusion in the message, and simultaneously argues against information redundancy---if the message were merely redundant information, reversing the conclusion should not change the outcome.

\paragraph{Testing the conclusion label (minimal-edit control).}
To further isolate the conclusion label, we construct minimal-edit messages: only the final answer line and one sentence stating the conclusion are changed, leaving all other reasoning intact.
Four of nine tested cells reach significance, confirming that the conclusion label alone has an independent causal effect on the receiver's answer, with supporting reasoning amplifying it.
Effect sizes vary by receiver and task.

\subsection{Narrowing Alternative Explanations}
\label{sec:results-rule-out}

The controls below test the four alternative explanations listed in the analysis framework.

\begin{figure*}[t]
  \centering
  \includegraphics[width=\textwidth]{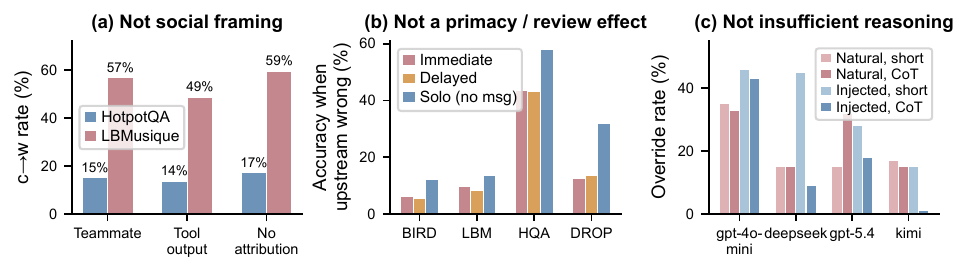}
  \caption{Source-label control, delayed-receipt control, and CoT override rates.}
  \label{fig:ruling-out}
\end{figure*}

\paragraph{Social framing does not account for the pattern.}
If substitution stems from social deference to a ``teammate'' label, changing the source label should change the following rate.
We relabel the message as an ``unverified tool output'' or remove attribution entirely on HotpotQA and LBMusique (Figure~\ref{fig:ruling-out}a).
The c$\to$w rates are statistically indistinguishable across conditions (all paired McNemar $p > 0.3$). For gpt-4o-mini and deepseek, TOST equivalence tests confirm differences within $\pm 5$\pp; for kimi, some comparisons are underpowered due to low base rates; on LBM, sample sizes are too small for conclusive equivalence testing (Appendix~\ref{sec:app-source-label}). On HotpotQA, where sample sizes are adequate, social framing does not drive the behavior.
Even an explicit instruction to prioritize evidence does not reduce the override.

\paragraph{Not explained by a second-attempt artifact.}
If an additional inference call randomly changes answers, the hidden branch should show similar churn even without an informative message.
We run a matched two-branch control: both branches use the same number of inference calls, but one sees the peer message and the other does not.
On both benchmarks (BIRD and HotpotQA), the message-exposed branch shows significantly more c$\to$w transitions than the matched control (BIRD gpt-4o-mini: 17 vs 2, $p < 0.001$; deepseek: 9 vs 0, $p = 0.004$).
Merely re-examining the answer does not produce substitution.

\paragraph{Strict primacy does not account for the pattern.}
Letting the receiver answer independently before seeing the message does not attenuate the effect (Figure~\ref{fig:ruling-out}b): on all four QA benchmarks, delayed receipt preserves the benefit of correct messages but fails to restore evidence use when the upstream answer is wrong.

\paragraph{Not explained by a scoring artifact.}
If format changes caused by the message inflate apparent harm, alternative scoring should give different conclusions.
Yes/No core extraction reduces but does not eliminate the override rate, and the affected benchmarks are limited to a specific subset (Appendix~\ref{sec:app-cot}).

\subsection{Recovery and Its Limits}
\label{sec:results-predict}

The recovery experiments below use oracle error knowledge to establish an upper bound on what targeted interventions can achieve; practical deployment requires imperfect detection, whose cost-benefit trade-off we analyze in \S\ref{sec:discussion}.

\begin{figure*}[t]
  \centering
  \includegraphics[width=\textwidth]{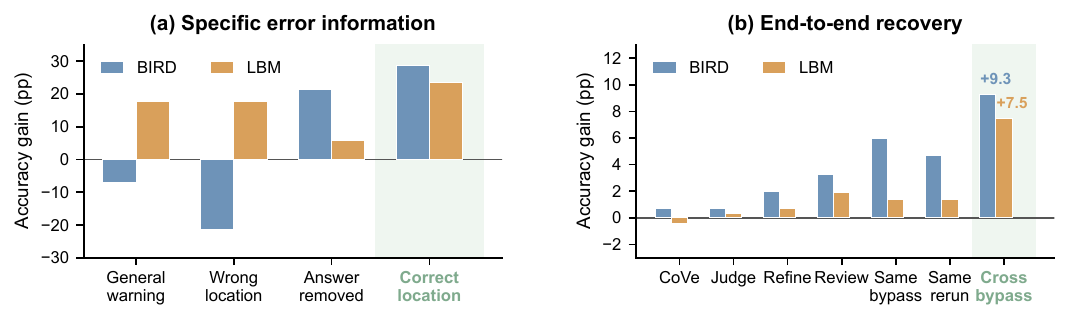}
  \caption{Error-location interventions and gated recovery actions.}
  \label{fig:intervention}
\end{figure*}

Starting from the same initial answer, we test five message variants: (1) original message (baseline), (2) generic warning, (3) incorrect error location, (4) upstream reasoning without the draft answer, and (5) correct error location.
Providing the correct error location yields the largest gains (Figure~\ref{fig:intervention}a; BIRD $p = 0.003$, LBM $p = 0.016$, item-level majority-vote McNemar).
Generic warnings yield smaller gains, and an \emph{incorrect} error location does not improve performance, confirming that the receiver acts on specifics, not mere ``be careful'' signals.

On upstream-wrong items, removing the message and regenerating with a different receiver yields significant recovery on all three benchmarks (Table~\ref{tab:cf-decomp-cond}).
Message removal alone is non-negative, and receiver replacement adds a further $+10.9$ to $+24.1$\pp; same-family and cross-family replacements achieve comparable gains (Appendix~\ref{sec:app-capability-matched}).
On upstream-correct items, message removal is harmful ($-20.7$ to $-25.4$\pp; Appendix Table~\ref{tab:cf-decomp-all}), so any deployment must gate the intervention on error detection.
The breakeven detector precision is approximately 66\% on BIRD and 94\% on L2W; details of the cost-benefit analysis appear in \S\ref{sec:discussion}.

\begin{table}[t]
\centering\small
\begin{tabular}{@{}lrrr@{}}
\toprule
Component & BIRD ($n{=}92$) & LBM ($n{=}70$) & L2W ($n{=}16$) \\
\midrule
A: Message removal     & $+10.9$ & $+6.7$  & $+1.7$  \\
B: Receiver replacement & $+10.9$ & $+24.1$ & $+22.3$ \\
C: Total               & $+21.7$ & $+30.8$ & $+24.0$ \\
\bottomrule
\end{tabular}
\caption{Three-way decomposition \textbf{on upstream-wrong items} (paired bootstrap, $B{=}10{,}000$). On upstream-correct items, message removal is harmful (Appendix Table~\ref{tab:cf-decomp-all}).}
\label{tab:cf-decomp-cond}
\end{table}

\section{Chain-of-Thought Diagnosis}
\label{sec:cot}

This section serves a dual purpose: it tests a natural defense (chain-of-thought reasoning) and uses the resulting traces as evidence for how the receiver processes conflicting information.
If substitution reflects over-weighting the peer's message, explicit step-by-step reasoning should help the receiver re-engage its own evidence.

CoT does not reliably eliminate the override.
On HotpotQA, two of four models show modest reductions in the override rate, while one shows a reversal and another also worsens; none reaches significance (Appendix~\ref{sec:app-cot}).
On LBM, the strongest reduction still leaves the model overriding nearly half of items it would answer correctly alone.
The picture is not ``CoT is useless'' but rather \emph{CoT is unreliable as a defense against naturally embedded errors}: it helps some models on some items, hurts others, and never eliminates the phenomenon.

The reasoning traces provide further evidence.
We classify override traces from four models and annotate 60 traces from the two stronger models (protocol in Appendix~\ref{sec:app-cot-protocol}).
Two independent annotators agree with near-perfect reliability that the receiver is \emph{evidence-engaged}: it cites relevant evidence yet follows the peer's wrong answer (prevalence-adjusted agreement AC1 $= 0.98$; Appendix~\ref{sec:app-cot}).
The failure is not for lack of evidence engagement.
This pattern is not predicted by a pure capacity account, which would expect the receiver to fail to engage evidence at all; whether it is consistent with anchoring or other accounts remains an open question.

\section{Related Work}
\label{sec:related}

\paragraph{Communication reliability in multi-agent LLM systems.}
Sequential multi-agent pipelines pass intermediate results from one agent to the next~\citep{ICLR2024_6507b115,wu2024autogen,zhang2024aflow,hu2024adas,chen2024agentverse}.
A growing body of work shows that errors propagate through such systems: \citet{cemri2025mast} taxonomize 14 failure modes; \citet{zhang2026hallucination} and \citet{guan2026snowball} show that hallucinations compound across stages; \citet{pan2026misinformation} study injected misinformation spread; and \citet{yang2026wrongbutuseful} show that wrong-answer messages can still carry useful intermediate steps.
These works establish \emph{that} errors propagate but do not ask \emph{why they persist when the downstream agent holds sufficient evidence to correct them}.
We study this question at the level of a single message handoff, the atomic unit of every sequential pipeline, and show that the effect is not mere propagation but a directed displacement toward the peer's specific wrong answer.

\paragraph{Evidence conflicts and social influence in LLMs.}
When an LLM receives conflicting inputs, several mechanisms can shift its answer: anchoring biases outputs toward prior numbers~\citep{ranaldi2024anchoring,wei2024anchorbench}, sycophancy toward user preferences~\citep{sharma2024sycophancy,perez2023discovering,wei2023sycophancy}, conformity toward simulated peer majorities~\citep{qu2026conformity,cho2025herd,shehata2026bystander}, and context-sensitivity studies show that irrelevant information~\citep{shi2023distraction} or positional biases~\citep{liu2024lost} can distract models.
The closest prior work is \citet{qu2026conformity}, who show that peer opinions induce conformity in multi-agent discussion.
Our setting differs in that the receiver holds independent evidence sufficient to answer correctly, and we manipulate the message while holding that evidence fixed, enabling causal attribution at the item level.
Whether the displacement we observe shares a mechanism with group conformity remains open.
Related work on knowledge conflicts~\citep{xie2024conflict,chen2022rich} studies contradictions between parametric and contextual knowledge; our conflict is between two external inputs, the peer message and the task evidence.

\paragraph{Self-correction and reasoning faithfulness.}
\citet{huang2024selfcorrect} show that without external feedback, LLM self-correction degrades performance.
Prompting techniques such as Self-Refine~\citep{madaan2024selfrefine} and Chain-of-Verification~\citep{dhuliawala2024cove} aim to catch errors through structured re-examination, yet \citet{huang2024selfcorrect} find their effectiveness is limited without external feedback.
Separately, chain-of-thought explanations are known to be unfaithful to the model's actual reasoning process~\citep{turpin2024unfaithful,lanham2023measuring,lyu2024faithful,chen2025reasoning}.
Our CoT analysis (\S\ref{sec:cot}) connects these two threads: when the receiver reasons step-by-step yet still follows the peer's wrong answer, the traces show that the receiver cites correct evidence yet adopts the peer's conclusion; the failure is not for lack of evidence engagement.

\section{Discussion and Conclusion}
\label{sec:discussion}

\paragraph{What substitution is, and what it does not establish.}
On items the receiver can solve alone, an erroneous peer message causes a directional shift toward the peer's specific wrong answer, and this shift is robust to timing and framing manipulations.
Our controlled experiments narrow the space of explanations but do not fully adjudicate between remaining accounts: the data do not distinguish whether the receiver internally replaces evidence-based reasoning with the peer's claim, or considers both inputs but assigns excessive weight to the message.
The trace-level finding that receivers cite correct evidence yet follow the peer's wrong answer is one that prior studies, which lack per-item evidence controls, could not have observed.

\paragraph{Implications for communication design.}
A common assumption in pipeline design is that passing intermediate results forward provides a free cross-check: if the upstream is right, the downstream benefits; if the upstream is wrong, the downstream can fall back on its own evidence.
Our results confirm the first half but challenge the second: the aggregate message value with evidence is close to zero (weighted mean $\tau = -0.6$\pp, 95\% CI $[-4.1, +3.3]$), because help and harm nearly cancel across items.
Prompting-based defenses do not reliably restore independent judgment (\S\ref{sec:cot},~\S\ref{sec:results-predict}), and all five tested receiver families exhibit the pattern on at least one benchmark.
Error-gated recovery (\S\ref{sec:results-predict}) can recoup losses when the upstream is wrong, but on the full item set message removal is net-harmful because discarding correct signals outweighs the benefit.
The breakeven detector precision---below which gated removal is net-harmful---is approximately 66\% on BIRD (where 61\% of upstream answers are wrong) and 94\% on L2W (13\% upstream errors), derived from Tables~\ref{tab:cf-decomp-cond} and~\ref{tab:cf-decomp-all}.
A regression of recovery on solo accuracy across eight replacement receivers ($R^2 = 0.77$; Appendix~\ref{sec:app-capability-matched}) confirms that capability explains most but not all of the variance; model diversity contributes beyond capability.
The practical implication is not that pipelines should stop passing messages, but that the message's value depends on the receiver's evidence state and on the upstream's correctness; both must be evaluated, not assumed.

\paragraph{Scope and open questions.}
Two conditions limit the generality of our findings.
First, all experiments use oracle-quality evidence (gold paragraphs, full schemas); in deployments where retrieval is imperfect, the incidence of substitution may differ from our measurements.
Second, our findings are demonstrated on tasks with discrete, verifiable answers; whether substitution extends to open-ended generation, iterative debate, or longer chains remains untested.
The central implication is that the benefit of communication coexists with a conditional cost: on items the receiver can solve independently, possessing evidence does not guarantee its effective use once an erroneous peer message enters the context.
How to preserve the benefits of communication while protecting independent judgment when the upstream is wrong remains an open problem that our controlled setting does not yet address.

\bibliography{bib/references}
\bibliographystyle{iclr2027_conference}

\newpage
\appendix
\section{Additional Results}
\label{sec:appendix}

\subsection{Benchmark Details}
\label{sec:app-benchmark-details}

\begin{table}[h]
\centering
\footnotesize
\setlength{\tabcolsep}{4pt}
\caption{Benchmark overview. ``Upstream err.''\ is the fraction of items where the upstream agent (gpt-4o-mini) produces an incorrect answer.}
\label{tab:hosts}
\begin{tabular}{@{}llrrll@{}}
\toprule
Benchmark & Task type & $n$ & Upstream err. & Indep.\ evidence & Metric \\
\midrule
BIRD~\citep{li2024bird} & SQL generation & 150 & 61.3\% & schema desc. & exec.\ acc. \\
L2W~\citep{ho2020l2w} & Multi-hop QA & 120 & 13.3\% & retrieved pass. & token F1 \\
LBM~\citep{trivedi2022musique} & Multi-hop QA & 160 & 43.8\% & retrieved pass. & token F1 \\
HotpotQA~\citep{yang2018hotpotqa} & Multi-hop QA & 200 & 50.0\% & gold paragraphs & token F1 \\
DROP~\citep{dua2019drop} & Reading comp. & 120 & 18.3\% & passage + question & exact match \\
\bottomrule
\end{tabular}
\end{table}

\subsection{Decomposition of kimi-k2.6 Upstream on LBM}
\label{sec:app-kimi-decomposition}

When kimi-k2.6 serves as upstream on LBM, its message value with independent evidence is $+10.5$\pp{} on items where the upstream answer is wrong, which appears to contradict the substitution pattern.
Splitting by receiver capability reveals the opposite:

\begin{table}[h]
\centering\small
\begin{tabular}{@{}lrrr@{}}
\toprule
Receiver can solve without message? & $n$ & \shortstack{Message value\\with evidence} & Interpretation \\
\midrule
Yes (score $> 0.5$) & 18 & $\mathbf{-22.8}$\pp & Substitution \\
No (score $\leq 0.5$) & 69 & $+19.1$\pp & Reasoning helps \\
\midrule
All upstream-wrong items & 87 & $+10.5$\pp & — \\
\bottomrule
\end{tabular}
\caption{LBM items where kimi-k2.6 gives an incorrect upstream answer. Positive message value with independent evidence comes from items the receiver cannot solve alone; on items it can solve, the wrong conclusion causes harm.}
\label{tab:kimi-decompose}
\end{table}

On the 22 items where the wrong message helps, 91\% (20/22) have receiver scores of zero without the peer message; the receiver cannot solve these multi-hop questions without guidance.
In 86\% of these items, the gold-standard answer appears within kimi's reasoning chain even though its final answer is wrong: kimi performs correct entity lookups and passage citations but arrives at an incorrect conclusion.
The receiver extracts useful intermediate steps from the reasoning, not the conclusion.

This decomposition is fully consistent with the substitution claim: the benefit comes from items the receiver would have failed alone, while the harm falls on items it would have solved alone.
A stronger upstream model provides higher-quality reasoning chains (kimi's mean reasoning length: 1,640 characters vs.\ gpt-4o-mini's $\approx$400), amplifying the benefit on hard items---but its wrong conclusions are equally capable of overriding the receiver's correct answers on easy ones.

\subsection{Per-Item Transfer Matrix (BIRD)}
\label{sec:app-transfer-matrix}

For the 92 BIRD items where the upstream answer is wrong (gpt-4o-mini receiver), we decompose message value with independent evidence into per-item outcomes:

\begin{itemize}[nosep]
  \item 11 items are \emph{harmed}: the receiver answers correctly without the message but incorrectly with it.
  \item 0 items are helped in the reverse direction.
  \item 81 items show no change (6 correct, 75 incorrect under both conditions).
\end{itemize}

\paragraph{Peer-answer adoption rate.}
\label{sec:app-adoption-rate}
Among correct-to-wrong transitions, we test whether the receiver adopts the peer's \emph{specific} wrong answer (token-F1 $\geq 0.5$ or substring containment) rather than producing an unrelated error.
The audit covers all c$\to$w items from four HotpotQA receivers (gpt-4o-mini, deepseek-v3.2, kimi-k2.6, gpt-5.4) and one LBM receiver (gpt-4o-mini)---82 items total, representing every c$\to$w transition in these five cells.
Of these, 77 (94\%, bootstrapped 95\% CI [87\%, 98\%]) show the receiver adopting the peer's specific wrong answer.
All four HQA receivers individually show adoption rates $\geq 83\%$.
As a chance baseline, open-ended QA answers are drawn from a large entity space; even conservatively assuming only 5 plausible wrong answers per item, random matching would yield $\sim$20\%, far below the observed 94\%.
This rules out the interpretation that the message merely confuses the receiver into random errors: the receiver copies the peer's specific conclusion.

\subsection{Behavioral Taxonomy}
\label{sec:app-taxonomy}

Table~\ref{tab:taxonomy} classifies every item where the upstream answer is wrong and the receiver has independent evidence.

\begin{table}[h]
\centering\small
\begin{tabular}{@{}lrrrr@{}}
\toprule
Behavior & BIRD & LBM & HQA & DROP \\
\midrule
Copies upstream error       & 84 & 66 & 38 & 86 \\
Ignores error, correct      &  7 &  4 & 37 &  0 \\
\textbf{Can solve, still follows} & \textbf{9} & \textbf{6} & \textbf{23} & \textbf{9} \\
Cannot solve, diverges      &  0 & 21 &  0 &  5 \\
Diverges, incorrect         &  0 &  3 &  0 &  0 \\
Diverges, correct           &  0 &  0 &  2 &  0 \\
\midrule
Follows upstream error      & 93 & 71 & 61 & 95 \\
\bottomrule
\end{tabular}
\caption{Behavioral classification for the gpt-4o-mini receiver (\%). Percentages are rounded independently; ``Follows upstream error'' combines unconditional copying with cases where the receiver can solve alone but still follows. Column sample sizes: BIRD 92, LBM 70, HQA 100, DROP 22.}
\label{tab:taxonomy}
\end{table}

Unconditional following accounts for 38--86\% of cases: the receiver copies the upstream error and cannot solve the item independently.
In a further 6--23\% of cases, the receiver can solve the item alone but still copies the upstream error when the message is present; this is the most direct evidence for substitution.

\subsection{BIRD-Hard Subset}
\label{sec:app-bird-hard}

BIRD-Hard uses the same pipeline and scoring as BIRD on a harder item subset ($n$=150, 68.0\% upstream errors).
The substitution pattern replicates: message value is $+28.0$ without independent evidence and $+1.3$ with independent evidence, giving $\Gamma = +26.7$.
The interaction is consistent with the main BIRD finding ($\Gamma_\text{BIRD} = +25.3$); evidence reduces message value from substantial to near-zero.

\subsection{L2W Results}
\label{sec:app-l2w}

L2W (LongBench-2WikiMQA) has only 16 items with a naturally incorrect upstream answer, limiting statistical power.
We report it for completeness: message value is $+46.6$ without independent evidence and $+21.8$ with independent evidence, giving $\Gamma = +24.8$ ($p < 0.0001$, $B = 10{,}000$ bootstrap draws).
In the conclusion-reversal experiment, correcting erroneous conclusions changes accuracy by $+46.0$ [25.0, 67.9], while corrupting correct conclusions changes it by $-72.1$ [$-81.4$, $-61.9$]; both effects are significant despite the small $n$.

\subsection{DROP Results}
\label{sec:app-drop}

DROP ($n$=120, 18.3\% upstream errors) uses exact-match scoring with a contains fallback for multi-span answers.
Table~\ref{tab:drop-four-cell} reports the four-cell decomposition for the primary receiver (gpt-4o-mini).
Table~\ref{tab:drop-cross-receiver} extends the analysis to all five receivers.

\begin{table}[h]
\centering\small
\begin{tabular}{@{}lrrrr@{}}
\toprule
Condition & Shown (\%) & Hidden (\%) & $\tau$\,pp & \\
\midrule
Without independent evidence & 86.0 & 14.2 & $+71.8$ & \\
With independent evidence    & 86.8 & 63.6 & $+23.2$ & \\
\midrule
$\Gamma$ & & & $+48.6$ & $p < 0.0001$ \\
\bottomrule
\end{tabular}
\caption{DROP four-cell decomposition (gpt-4o-mini receiver, $n$=120). $\tau$ = message value (shown $-$ hidden accuracy). $\Gamma = \tau(\text{without evidence}) - \tau(\text{with evidence})$. DROP shows the largest $\Gamma$ among all five benchmarks. Unlike HotpotQA and LBM, $\tau(\text{with evidence})$ remains positive ($+23.2$\,pp): the message is net-helpful even with evidence. Independent evidence sharply reduces the marginal value of the message.}
\label{tab:drop-four-cell}
\end{table}

Note that $\tau(\text{with evidence}) = +23.2$\,pp remains positive: the peer message is net-helpful even when the receiver has independent evidence.
The large positive $\Gamma$ confirms that evidence reduces the marginal value of the message, consistent with the other four benchmarks.


\begin{table}[h]
\centering\footnotesize
\setlength{\tabcolsep}{4pt}
\begin{tabular}{@{}lrrrrrrrr@{}}
\toprule
Receiver & \multicolumn{2}{c}{With evidence} & \multicolumn{2}{c}{No evidence} & $\tau$(w/o\,ev.) & $\tau$(w/\,ev.) & $\Gamma$ [95\% CI] & c$\to$w \\
\cmidrule(lr){2-3}\cmidrule(lr){4-5}
 & Shown & Hidden & Shown & Hidden & & & & \\
\midrule
gpt-4o-mini   & 86.8 & 63.6 & 86.0 & 14.2 & $+71.8$ & $+23.2$ & $+48.6$              & 5 \\
deepseek-v3.2 & 90.8 & 70.8 & 90.0 & 14.2 & $+75.8$ & $+20.0$ & $+55.8$ {\scriptsize[45,66]} & 3 \\
kimi-k2.6     & 90.8 & 83.3 & 90.8 & 21.7 & $+69.2$ & $+7.5$  & $+61.7$ {\scriptsize[52,71]} & 7 \\
glm-5         & 92.5 & 90.8 & 90.8 & 20.0 & $+70.8$ & $+1.7$  & $+69.2$ {\scriptsize[61,78]} & 2 \\
qwen3.6-plus  & 92.5 & 94.2 & 90.0 & 24.2 & $+65.8$ & $-1.7$  & $+67.5$ {\scriptsize[58,76]} & 6 \\
\bottomrule
\end{tabular}
\caption{DROP cross-receiver four-cell decomposition ($n$=120 per receiver). Accuracy is exact-match percentage. $\Gamma$ is significantly positive for all five receivers ($p < 0.0001$); 95\% CIs from paired bootstrap ($B$=10,000).}
\label{tab:drop-cross-receiver}
\end{table}

\subsection{Recovery Experiment: Intervention Comparison}
\label{sec:app-recovery}

To test whether substitution can be overcome by known repair strategies, we apply a shared error detector (gemini-2.5-pro) and route flagged items to different interventions.
The detector is prompted with the upstream message and the receiver's evidence and asked whether the upstream conclusion is likely wrong.
Table~\ref{tab:detector-diag} reports its operating characteristics.

\begin{table}[h]
\centering\small
\begin{tabular}{@{}lccc@{}}
\toprule
Benchmark & Precision (\%) & Recall (\%) & Items flagged \\
\midrule
BIRD ($n$=150) & 91.8 & 60.9 & 61 \\
LBM ($n$=160) & 81.1 & 42.9 & 37 \\
L2W ($n$=120) & 35.0 & 43.8 & 20 \\
\bottomrule
\end{tabular}
\caption{Error-detector operating characteristics (gemini-2.5-pro). Precision = fraction of flagged items that are truly upstream-wrong; recall = fraction of upstream-wrong items that are flagged. L2W has low precision because most upstream answers are correct (86.7\%), so even modest false-positive rates dominate.}
\label{tab:detector-diag}
\end{table}

Table~\ref{tab:recovery} reports gated accuracy: unflagged items retain the original answer; flagged items receive the intervention.

\begin{table}[h]
\centering\small
\begin{tabular}{@{}lcccccc@{}}
\toprule
\multirow{2}{*}{Intervention} & \multicolumn{2}{c}{BIRD ($n$=150, fl.=61)} & \multicolumn{2}{c}{LBM ($n$=160, fl.=37)} & \multicolumn{2}{c}{L2W ($n$=120, fl.=20)} \\
\cmidrule(lr){2-3}\cmidrule(lr){4-5}\cmidrule(lr){6-7}
 & Acc & 95\% CI & Acc & 95\% CI & Acc & 95\% CI \\
\midrule
Default (no detection) & .400 & [.320, .480] & .539 & [.467, .609] & .742 & [.671, .807] \\
No message (same receiver) & .460 & [.380, .540] & .553 & [.482, .624] & .730 & [.656, .801] \\
Rerun (same receiver)      & .447 & [.367, .527] & .553 & [.481, .623] & .755 & [.686, .819] \\
CoVe                   & .407 & [.327, .487] & .535 & [.462, .606] & .746 & [.675, .813] \\
LLM-judge repair       & .407 & [.327, .487] & .542 & [.470, .614] & .746 & [.676, .813] \\
Self-refine            & .420 & [.340, .500] & .546 & [.474, .615] & .761 & [.693, .826] \\
Answer, then review message & .433 & [.353, .513] & .558 & [.487, .628] & .759 & [.690, .820] \\
\textbf{No message (different family)} & \textbf{.493} & [.413, .573] & \textbf{.614} & [.546, .681] & .769 & [.701, .833] \\
\bottomrule
\end{tabular}
\caption{Recovery experiment: gated accuracy under different interventions with gemini-2.5-pro as the shared detector. ``No message (different family)'' uses a receiver from another model family without the upstream message. ``Answer, then review message'' lets the receiver answer independently before reviewing the peer message.}
\label{tab:recovery}
\end{table}

\paragraph{Reviewing the message does not significantly improve over hiding it.}
Answering independently before reviewing the message and using the same receiver without the message produce statistically indistinguishable results on all three benchmarks ($\Delta = -2.7$, $+0.5$, $+2.9$\pp; all CIs contain zero).
Thus, ``think first, then review the message'' does not significantly improve over simply ignoring the message.
Using a different-family receiver without the message outperforms answer-then-review on BIRD by $+6.0$\pp{}; this gain persists even when the replacement receiver has matched solo capability (Table~\ref{tab:capability-matched}), jointly changing the receiver model and removing the message; we do not attribute the gain specifically to family diversity, as the replacement simultaneously changes capability, training distribution, and output style.

\paragraph{Summary of gated recovery results.}
Generic warnings (CoVe~\citep{dhuliawala2024cove}, LLM-judge~\citep{zheng2024judging}) are ineffective ($\leq +0.7$\pp{} on BIRD), and having the receiver deliberate before seeing the message does not restore verification.
Under the gated protocol (detector flags \textrightarrow{} intervention on flagged items only), the largest gains come from removing the message and using a different-family receiver ($+9.3$\pp{} on BIRD, $+7.5$\pp{} on LBM).
The conditional decomposition on upstream-wrong items (Table~\ref{tab:cf-decomp-cond} in the main text) shows that message removal is non-negative on all three benchmarks and that receiver replacement adds a substantial further gain.

\paragraph{Unconditional decomposition (all with-evidence items).}
For completeness, Table~\ref{tab:cf-decomp-all} reports the same decomposition on all with-evidence items regardless of upstream correctness.
Component~A is negative on LBM ($-7.7$\pp) and L2W ($-21.8$\pp): removing the message discards correct signals alongside erroneous ones, and the cost dominates when most upstream answers are correct (LBM: 56.2\%, L2W: 86.7\%).
This is the expected cost--benefit trade-off of indiscriminate message removal and does not contradict the conditional results: on the target population (upstream-wrong items), message removal does not hurt (Table~\ref{tab:cf-decomp-cond}).

\begin{table}[h]
\centering\small
\begin{tabular}{@{}lrrr@{}}
\toprule
Component & BIRD ($n{=}150$) & LBM ($n{=}160$) & L2W ($n{=}120$) \\
\midrule
A: Message removal     & $-1.3$  & $-7.7$  & $-21.8$ \\
B: Receiver replacement & $+9.7$  & $+17.4$ & $+14.8$ \\
C: Total               & $+8.3$  & $+9.7$  & $-7.0$  \\
\bottomrule
\end{tabular}
\caption{Three-way decomposition on \textbf{all with-evidence items} (unconditional on upstream correctness; paired bootstrap, $B{=}10{,}000$). Component~A is negative when most upstream answers are correct, because removing the message discards correct signals. On the upstream-wrong target population, Component~A is non-negative on all three benchmarks (Table~\ref{tab:cf-decomp-cond}). $C = A + B$ by construction.}
\label{tab:cf-decomp-all}
\end{table}

\paragraph{Capability-matched recovery and same-family control.}
\label{sec:app-capability-matched}

The cross-family receivers used in the main decomposition (kimi-k2.6, deepseek-v3.2) have higher solo accuracy than gpt-4o-mini on BIRD (49.3\% and 40.7\% vs.\ 43.3\%), raising the concern that the recovery benefit reflects capability differences rather than model replacement per se.
To address this, we evaluate eight replacement receivers on the same 92 upstream-wrong BIRD items and report each receiver's solo accuracy alongside its recovery effect (Table~\ref{tab:capability-matched}).
Crucially, the eight replacement receivers include two \textbf{same-family} OpenAI models (gpt-4o and gpt-4.1-mini) alongside six cross-family receivers, enabling a direct test of whether recovery requires \emph{family} diversity or merely a \emph{different model}.

\textbf{Same-family result.}
gpt-4.1-mini (solo 44.0\%, $\Delta = +0.7$\,pp vs.\ gpt-4o-mini) is essentially capability-matched and yields $+17.4$\,pp recovery ($p < 0.001$), comparable to the best cross-family receivers.
Across the 92 upstream-wrong BIRD items, the per-item mean of the two same-family replacements (25.0\%) is statistically indistinguishable from the per-item mean of the two cross-family replacements in the main decomposition (27.2\%; paired $t$-test $p = 0.50$, bootstrap 95\% CI of the difference $[-4.3, +8.7]$\,pp).
The same pattern holds on L2W ($+8.9$\,pp gap, $p = 0.21$).
On LBM, cross-family receivers outperform same-family ones by $+27.4$\,pp ($p < 0.001$), but this gap is explained by a capability confound: gpt-4o and gpt-4.1-mini are \emph{weaker} than gpt-4o-mini on LBM solo (41.9\% and 42.2\% vs.\ 46.2\%), while the cross-family receivers are substantially stronger (qwen3.6-plus 72.6\%, glm-5 71.9\%, kimi-k2.6 65.6\%).

We therefore conclude that the receiver-replacement benefit on BIRD is driven by using a \emph{different model}---not specifically a different model family.
This is consistent with the reframing in the main text: we label Component~B ``receiver replacement'' rather than ``family diversity'' and do not claim a causal role for family membership.

\textbf{Capability--recovery regression.}
To quantify how much of the recovery benefit is explained by capability differences, we regress recovery on solo accuracy across the eight replacement receivers (excluding the gpt-4o-mini baseline).
Solo accuracy explains most of the recovery variance ($R^2 = 0.77$, $p = 0.002$, slope $= 1.17 \pm 0.24$\pp per percentage point of solo accuracy). The residual variance is consistent with per-item model complementarity, but could also reflect unmeasured confounds; our design does not isolate model diversity as a causal factor.

\begin{table}[h]
\centering\small
\begin{tabular}{@{}llrrrl@{}}
\toprule
Receiver & Family & Solo acc.\ (\%) & $\Delta$ vs gpt (\pp) & Recovery (\pp) & $p$ \\
\midrule
\multicolumn{6}{@{}l}{\emph{Same-family replacements (OpenAI):}} \\
gpt-4.1-mini & OpenAI & 44.0 & $+0.7$ & $+17.4$ & $< 0.001$ \\
gpt-4o & OpenAI & 48.7 & $+5.4$ & $+21.7$ & $< 0.001$ \\
\midrule
\multicolumn{6}{@{}l}{\emph{Cross-family replacements (capability-matched, $|\Delta| \leq 5$\,pp):}} \\
deepseek-v3.2 & DeepSeek & 40.7 & $-2.7$ & $+9.8$ & $0.022$ \\
glm-4.5-air & Zhipu & 42.0 & $-1.3$ & $+17.4$ & $< 0.001$ \\
MiniMax-M2.5 & MiniMax & 46.7 & $+3.3$ & $+14.1$ & $< 0.001$ \\
\midrule
\multicolumn{6}{@{}l}{\emph{Cross-family replacements (stronger):}} \\
kimi-k2.6 & Moonshot & 49.3 & $+6.0$ & $+17.4$ & $< 0.001$ \\
qwen-plus & Alibaba & 56.0 & $+12.7$ & $+31.5$ & $< 0.001$ \\
gemini-2.5-flash-lite & Google & 57.3 & $+14.0$ & $+28.3$ & $< 0.001$ \\
\bottomrule
\end{tabular}
\caption{Recovery by receiver identity on BIRD ($n$=92 upstream-wrong items, paired bootstrap).
Solo accuracy is computed on 150 BIRD items without any peer message in an independent run.
Recovery = paired accuracy gain over the default gpt-4o-mini receiver with the erroneous message; the default receiver's same-model removal baseline is $+10.9$\pp (Table~\ref{tab:cf-decomp-cond}).
Same-family OpenAI replacements achieve recovery comparable to cross-family receivers at similar capability levels ($p = 0.50$ for same-family vs.\ cross-family mean difference), indicating that recovery does not require family diversity---any different model suffices.}
\label{tab:capability-matched}
\end{table}

\subsection{Message Value and Interaction (Full 20-Cell Table)}
\label{sec:app-gamma}

Table~\ref{tab:gamma-20} extends the single-receiver $\Gamma$ values from the main text to all 20 receiver--benchmark cells.
In all 20 cells, $\Gamma$ is large and significant ($p < 0.0001$): independent evidence sharply reduces the marginal value of the upstream message.

\begin{table}[h]
\centering\small
\begin{tabular}{@{}llrrrrr@{}}
\toprule
Receiver & Bench & $n$ & $\tau$(w/o\,ev.)\,pp & $\tau$(w/\,ev.)\,pp & $\Gamma$\,pp & 95\% CI \\
\midrule
gpt-4o-mini   & BIRD & 150 & $+26.7$ & $+1.3$   & $+25.3$ & $[+17.3, +33.3]$ \\
gpt-4o-mini   & LBM  & 160 & $+37.9$ & $+7.7$   & $+30.2$ & $[+22.1, +38.6]$ \\
gpt-4o-mini   & L2W  & 120 & $+46.6$ & $+21.8$  & $+24.8$ & $[+15.2, +34.4]$ \\
gpt-4o-mini   & HQA  & 200 & $+34.3$ & $-5.7$   & $+40.0$ & $[+33.7, +46.6]$ \\
\midrule
deepseek-v3.2 & BIRD & 150 & $+27.3$ & $-4.0$   & $+31.3$ & $[+22.7, +40.0]$ \\
deepseek-v3.2 & LBM  & 160 & $+38.3$ & $+10.5$  & $+27.8$ & $[+20.7, +34.8]$ \\
deepseek-v3.2 & L2W  & 120 & $+43.2$ & $+16.8$  & $+26.4$ & $[+16.8, +36.0]$ \\
deepseek-v3.2 & HQA  & 200 & $+32.8$ & $-5.2$   & $+38.0$ & $[+31.3, +44.9]$ \\
\midrule
kimi-k2.6     & BIRD & 150 & $+16.7$ & $-12.0$  & $+28.7$ & $[+20.0, +38.0]$ \\
kimi-k2.6     & LBM  & 160 & $+17.4$ & $-10.2$  & $+27.6$ & $[+19.9, +35.5]$ \\
kimi-k2.6     & L2W  & 120 & $+15.6$ & $-7.1$   & $+22.7$ & $[+14.3, +31.0]$ \\
kimi-k2.6     & HQA  & 200 & $+30.4$ & $-5.8$   & $+36.2$ & $[+29.7, +42.6]$ \\
\midrule
glm-5         & BIRD & 150 & $+14.0$ & $-12.7$  & $+26.7$ & $[+18.7, +34.7]$ \\
glm-5         & LBM  & 160 & $+14.6$ & $-5.9$   & $+20.5$ & $[+13.5, +27.7]$ \\
glm-5         & L2W  & 120 & $+10.9$ & $-7.8$   & $+18.7$ & $[+10.4, +27.4]$ \\
glm-5         & HQA  & 200 & $+13.6$ & $-3.4$   & $+16.9$ & $[+11.3, +22.6]$ \\
\midrule
qwen3.6-plus  & BIRD & 150 & $+27.3$ & $-4.7$   & $+32.0$ & $[+22.7, +41.3]$ \\
qwen3.6-plus  & LBM  & 160 & $+16.3$ & $-5.4$   & $+21.7$ & $[+14.0, +29.2]$ \\
qwen3.6-plus  & L2W  & 120 & $+12.8$ & $-6.1$   & $+18.9$ & $[+10.4, +27.7]$ \\
qwen3.6-plus  & HQA  & 200 & $+20.0$ & $-6.3$   & $+26.4$ & $[+20.9, +32.1]$ \\
\bottomrule
\end{tabular}
\caption{Full 20-cell $\Gamma$ table. $\tau$(w/o\,ev.) = message value without independent evidence; $\tau$(w/\,ev.) = message value with evidence; $\Gamma = \tau\text{(w/o\,ev.)} - \tau\text{(w/\,ev.)}$. All CIs from paired bootstrap ($B=10{,}000$, seed~42). All 20 cells show $\Gamma > 0$ ($p < 0.0001$): independent evidence consistently reduces the marginal value of the upstream message.}
\label{tab:gamma-20}
\end{table}

\subsection{Per-Item Bootstrap CIs for $\tau$(with evidence)}
\label{sec:app-tau-ci}

Table~\ref{tab:tau-ci} reports paired bootstrap 95\% CIs for $\tau(\text{with evidence})$ on all 20 receiver--benchmark cells.

\begin{table}[h]
\centering\small
\begin{tabular}{@{}llrrrrl@{}}
\toprule
Receiver & Benchmark & $n$ & $\tau$(w/\,ev.)\,pp & 95\% CI & $p$ & \\
\midrule
glm-5            & BIRD & 150 & $-12.7$ & $[-19.3, -6.7]$ & ${<}.001$ & $\ast\ast\ast$ \\
kimi-k2.6        & BIRD & 150 & $-12.0$ & $[-18.7, -5.3]$ & ${<}.001$ & $\ast\ast\ast$ \\
qwen3.6-plus     & BIRD & 150 & $-4.7$  & $[-11.3, +2.0]$ & .191 &  \\
deepseek-v3.2    & BIRD & 150 & $-4.0$  & $[-11.3, +3.3]$ & .319 &  \\
gpt-4o-mini      & BIRD & 150 & $+1.3$  & $[-5.3, +8.0]$  & .769 &  \\
\midrule
kimi-k2.6        & LBM  & 160 & $-10.2$ & $[-16.0, -4.4]$ & ${<}.001$ & $\ast\ast\ast$ \\
glm-5            & LBM  & 160 & $-5.9$  & $[-10.3, -1.7]$ & .007 & $\ast\ast$ \\
qwen3.6-plus     & LBM  & 160 & $-5.4$  & $[-10.2, -0.9]$ & .019 & $\ast$ \\
gpt-4o-mini      & LBM  & 160 & $+7.7$  & $[+1.3, +14.2]$ & .018 & $\ast$ \\
deepseek-v3.2    & LBM  & 160 & $+10.5$ & $[+3.6, +17.3]$ & .002 & $\ast\ast$ \\
\midrule
qwen3.6-plus     & HQA  & 200 & $-6.3$  & $[-9.7, -3.3]$  & ${<}.001$ & $\ast\ast\ast$ \\
kimi-k2.6        & HQA  & 200 & $-5.8$  & $[-9.8, -2.0]$  & .002 & $\ast\ast$ \\
gpt-4o-mini      & HQA  & 200 & $-5.7$  & $[-9.5, -1.8]$  & .005 & $\ast\ast$ \\
deepseek-v3.2    & HQA  & 200 & $-5.2$  & $[-9.2, -1.3]$  & .007 & $\ast\ast$ \\
glm-5            & HQA  & 200 & $-3.4$  & $[-6.3, -0.7]$  & .013 & $\ast$ \\
\midrule
glm-5            & L2W  & 120 & $-7.8$  & $[-12.8, -3.2]$ & ${<}.001$ & $\ast\ast\ast$ \\
kimi-k2.6        & L2W  & 120 & $-7.1$  & $[-12.2, -2.6]$ & .004 & $\ast\ast$ \\
qwen3.6-plus     & L2W  & 120 & $-6.1$  & $[-10.9, -1.4]$ & .010 & $\ast$ \\
deepseek-v3.2    & L2W  & 120 & $+16.8$ & $[+9.0, +24.9]$ & ${<}.001$ & $\ast\ast\ast$ \\
gpt-4o-mini      & L2W  & 120 & $+21.8$ & $[+14.0, +30.0]$ & ${<}.001$ & $\ast\ast\ast$ \\
\bottomrule
\end{tabular}
\caption{Paired bootstrap 95\% CIs for $\tau$(with evidence) ($B = 10{,}000$ resamples, seed~42). Of 20 cells, 13 are significantly negative at $\alpha = 0.05$; 6 survive Bonferroni correction ($\alpha = 0.0025$ for 20 tests). $\ast$\,$p < .05$; $\ast\ast$\,$p < .01$; $\ast\ast\ast$\,$p < .001$.}
\label{tab:tau-ci}
\end{table}

Table~\ref{tab:transitions} decomposes $\tau(\text{with evidence})$ into per-item transitions for all 20 cells.

\begin{table}[h]
\centering\small
\begin{tabular}{@{}llrrrr@{}}
\toprule
Receiver & Bench & {Bypass correct} & \shortstack{$P(\text{c}{\to}\text{w})$} & {Bypass wrong} & \shortstack{$P(\text{w}{\to}\text{c})$} \\
\midrule
qwen3.6-plus  & BIRD &  96 & 16.7\% &  54 & 16.7\% \\
glm-5         & BIRD &  87 & 26.4\% &  63 &  6.3\% \\
kimi-k2.6     & BIRD &  75 & 32.0\% &  75 &  8.0\% \\
deepseek-v3.2 & BIRD &  70 & 25.7\% &  80 & 15.0\% \\
gpt-4o-mini   & BIRD &  58 & 20.7\% &  92 & 15.2\% \\
\midrule
qwen3.6-plus  & LBM  & 121 & 10.7\% &  39 & 12.8\% \\
glm-5         & LBM  & 120 & 10.0\% &  40 & 10.0\% \\
kimi-k2.6     & LBM  & 109 & 22.0\% &  51 & 21.6\% \\
gpt-4o-mini   & LBM  &  77 & 11.7\% &  83 & 27.7\% \\
deepseek-v3.2 & LBM  &  76 & 14.5\% &  84 & 33.3\% \\
\midrule
qwen3.6-plus  & HQA  & 180 &  8.3\% &  20 &  0.0\% \\
glm-5         & HQA  & 176 &  5.7\% &  24 &  8.3\% \\
deepseek-v3.2 & HQA  & 172 & 10.5\% &  28 & 25.0\% \\
kimi-k2.6     & HQA  & 172 & 11.0\% &  28 & 25.0\% \\
gpt-4o-mini   & HQA  & 159 & 13.2\% &  41 & 19.5\% \\
\midrule
glm-5         & L2W  & 100 &  9.0\% &  20 & 10.0\% \\
qwen3.6-plus  & L2W  & 100 &  7.0\% &  20 & 15.0\% \\
kimi-k2.6     & L2W  &  99 &  9.1\% &  21 & 14.3\% \\
deepseek-v3.2 & L2W  &  70 &  4.3\% &  50 & 50.0\% \\
gpt-4o-mini   & L2W  &  67 &  1.5\% &  53 & 54.7\% \\
\bottomrule
\end{tabular}
\caption{Per-item transitions under $\tau$(with evidence). $P(\text{c}{\to}\text{w})$: fraction of items correct without the message that become wrong with it. $P(\text{w}{\to}\text{c})$: the reverse. On HQA, all five receivers show non-trivial correct-to-wrong rates (5.7--13.2\%).}
\label{tab:transitions}
\end{table}

\subsection{Conditional Decomposition of $\tau$(with evidence)}
\label{sec:app-conditional}

Table~\ref{tab:conditional-decomp} decomposes $\tau$(with evidence) by receiver capability.
Items where the receiver answers correctly without the message ($n_{\rm solved}$) can only lose accuracy; items where it fails ($n_{\rm failed}$) can only gain.
The key quantity is $P(\text{c}{\to}\text{w})$: the fraction of independently solvable items where the message causes a wrong answer.

\begin{table}[h]
\centering\small
\begin{tabular}{@{}llrrrrrr@{}}
\toprule
Receiver & Bench & $n_s$ & c$\to$w & $\tau_s$\,pp & 95\% CI & $n_f$ & w$\to$c \\
\midrule
gpt-4o-mini   & BIRD &  58 & 12 & $-20.7$ & $[-31.0, -10.3]$ &  92 & 14 \\
gpt-4o-mini   & LBM  &  77 &  9 & $-11.7$ & $[-19.5, -5.2]$  &  83 & 23 \\
gpt-4o-mini   & L2W  &  67 &  1 & $-1.5$  & $[-4.5, 0.0]$    &  53 & 29 \\
gpt-4o-mini   & HQA  & 159 & 21 & $-13.2$ & $[-18.9, -8.2]$  &  41 &  8 \\
\midrule
deepseek-v3.2 & BIRD &  70 & 18 & $-25.7$ & $[-35.7, -15.7]$ &  80 & 12 \\
deepseek-v3.2 & LBM  &  76 & 11 & $-14.5$ & $[-22.4, -6.6]$  &  84 & 28 \\
deepseek-v3.2 & L2W  &  70 &  3 & $-4.3$  & $[-8.6, 0.0]$    &  50 & 25 \\
deepseek-v3.2 & HQA  & 172 & 18 & $-10.5$ & $[-15.7, -5.8]$  &  28 &  7 \\
\midrule
kimi-k2.6     & BIRD &  75 & 24 & $-32.0$ & $[-42.7, -21.3]$ &  75 &  6 \\
kimi-k2.6     & LBM  & 109 & 24 & $-22.0$ & $[-29.4, -14.7]$ &  51 & 11 \\
kimi-k2.6     & L2W  &  99 &  9 & $-9.1$  & $[-14.1, -4.0]$  &  21 &  3 \\
kimi-k2.6     & HQA  & 172 & 19 & $-11.0$ & $[-16.3, -6.4]$  &  28 &  7 \\
\midrule
glm-5         & BIRD &  87 & 23 & $-26.4$ & $[-35.6, -17.2]$ &  63 &  4 \\
glm-5         & LBM  & 120 & 12 & $-10.0$ & $[-15.0, -5.0]$  &  40 &  4 \\
glm-5         & L2W  & 100 &  9 & $-9.0$  & $[-14.0, -4.0]$  &  20 &  2 \\
glm-5         & HQA  & 176 & 10 & $-5.7$  & $[-9.7, -2.3]$   &  24 &  2 \\
\midrule
qwen3.6-plus  & BIRD &  96 & 16 & $-16.7$ & $[-24.0, -9.4]$  &  54 &  9 \\
qwen3.6-plus  & LBM  & 121 & 13 & $-10.7$ & $[-16.5, -5.8]$  &  39 &  5 \\
qwen3.6-plus  & L2W  & 100 &  7 & $-7.0$  & $[-12.0, -2.0]$  &  20 &  3 \\
qwen3.6-plus  & HQA  & 180 & 15 & $-8.3$  & $[-12.8, -4.4]$  &  20 &  0 \\
\bottomrule
\end{tabular}
\caption{Conditional decomposition of $\tau$(with evidence). $n_s$ = items solvable without message ($k{=}3$ majority vote); $\tau_s$ = message value on solvable items (always $\leq 0$). CIs from bootstrap ($B=10{,}000$). All c$\to$w transitions across the primary 20 cells: 274 out of 2{,}184 solvable items (12.5\% overall). Including DROP, the total is 297/2{,}667 (11.1\%). In 18 of 20 primary cells, the c$\to$w rate exceeds 5\%.}
\label{tab:conditional-decomp}
\end{table}

\subsection{Upstream Model Diversity}
\label{sec:app-upstream-diversity}

Table~\ref{tab:upstream-diversity-app} reports $\tau(\text{w/\,ev.})$ (message value with evidence) and $\Gamma$ for each upstream--benchmark combination.
All eight $\Gamma$ values are significant ($p < 0.0001$), confirming that evidence sharply reduces message value regardless of which model generates the message.
$\tau(\text{w/\,ev.})$ is non-negative in all eight cells: the message remains net-helpful on average even with evidence.
The substitution effect (high $\Gamma$) coexists with near-zero aggregate message value when evidence is present---it manifests as help and harm nearly canceling when the upstream error rate is moderate.
This means that unconditionally removing the message is not guaranteed to improve system performance; targeted interventions that preserve correct messages while mitigating harmful ones remain an open problem.

\begin{table}[h]
\centering
\small
\caption{Upstream model diversity: message value with evidence and interaction. $\tau$(w/\,ev.) = message value with evidence; $\Gamma = \tau$(w/o\,ev.)$\,-\,\tau$(w/\,ev.). All $\Gamma$ values significant ($p < 0.0001$). $\tau$(w/\,ev.) is non-negative in all cells.}
\label{tab:upstream-diversity-app}
\begin{tabular}{@{}llrrrr@{}}
\toprule
Upstream & Benchmark & Ups.\ corr. & $\tau$(w/\,ev.)\,pp & 95\% CI & $\Gamma$\,pp \\
\midrule
gpt-4o-mini & BIRD & 38.7\% & $+1.3$ & $[-5.3, +8.0]$ & $+25.3$ \\
gpt-4o-mini & LBM  & 56.2\% & $+7.7$ & $[+1.3, +14.2]$ & $+30.2$ \\
gpt-5.4 & HQA & 84.5\% & $+2.3$ & $[-1.1, +5.9]$ & $+43.1$ \\
gpt-5.4 & LBM      & 66.9\% & $+29.3$ & $[+21.8, +36.9]$ & $+21.4$ \\
kimi-k2.6 & HQA & 76.0\% & $+1.0$ & $[-2.5, +4.5]$ & $+42.6$ \\
kimi-k2.6 & LBM      & 45.6\% & $+20.4$ & $[+13.5, +27.3]$ & $+22.5$ \\
qwen-plus & HQA & 73.0\% & $+4.4$ & $[+0.4, +8.7]$ & $+37.6$ \\
qwen-plus & LBM      & 37.5\% & $+16.1$ & $[+9.6, +22.6]$ & $+20.1$ \\
\bottomrule
\end{tabular}
\end{table}

\subsection{Conclusion-Reversal Experiment: Full Effect Sizes}
\label{sec:app-flip-ci}

\begin{figure}[h]
  \centering
  \includegraphics[width=0.75\linewidth]{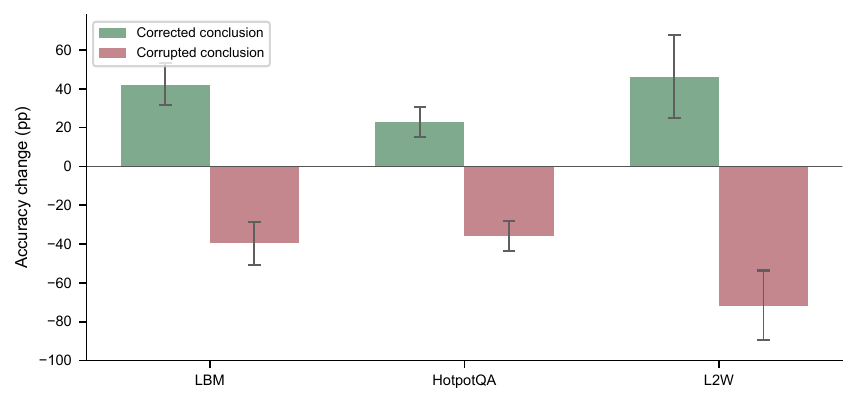}
  \caption{Conclusion-reversal experiment on items with an incorrect upstream answer (three QA benchmarks). Points show the accuracy difference between messages with corrected and original peer conclusions, with 95\% bootstrap CIs. All six contrasts are significant (BIRD and DROP omitted; see text). A compact version appears in the main text as Figure~\ref{fig:ruling-out}(a).}
  \label{fig:app-flip}
\end{figure}

\begin{table}[h]
\centering\small
\begin{tabular}{@{}l rr rr@{}}
\toprule
 & \multicolumn{2}{c}{Upstream wrong} & \multicolumn{2}{c}{Upstream correct} \\
\cmidrule(lr){2-3}\cmidrule(lr){4-5}
Benchmark & $\Delta$ & 95\% CI & $\Delta$ & 95\% CI \\
\midrule
LBM      & $+42.2$ & $[31.6,\;53.1]$     & $-39.4$ & $[-49.3,\;-29.6]$ \\
HotpotQA & $+22.7$ & $[15.1,\;30.4]$     & $-35.9$ & $[-44.8,\;-27.6]$ \\
L2W      & $+46.0$ & $[25.0,\;67.9]$     & $-72.1$ & $[-81.4,\;-61.9]$ \\
\bottomrule
\end{tabular}
\caption{Effects of reversing the peer conclusion, with 95\% bootstrap CIs (20K draws). All six contrasts on the three QA benchmarks are significant (BIRD and DROP rows omitted due to SQL-execution scoring and source-wrong label divergence; raw values retained in comments). ``Upstream wrong'': corrected $-$ original; ``Upstream correct'': corrupted $-$ original.}
\label{tab:flip-ci}
\end{table}

\subsection{Complete Four-Cell Accuracy Table}
\label{sec:app-four-cell}

Table~\ref{tab:four-cell} reports the four cell means underlying the $\tau$ and $\Gamma$ estimates for the primary receiver (gpt-4o-mini).
Cell means are computed from paired per-item scores.
DROP four-cell decompositions for all five receivers are in Appendix~\ref{sec:app-drop}.

\begin{table}[h]
\centering\small
\begin{tabular}{@{}l rrr rrr@{}}
\toprule
 & \multicolumn{3}{c}{Without evidence} & \multicolumn{3}{c}{With evidence} \\
\cmidrule(lr){2-4}\cmidrule(lr){5-7}
Benchmark & Shown & Hidden & $\tau$ & Shown & Hidden & $\tau$ \\
\midrule
BIRD & 31.3 &  4.7 & $+26.7$ & 40.0 & 38.7 & $+1.3$ \\
LBM  & 52.5 & 14.6 & $+37.9$ & 53.9 & 46.2 & $+7.7$ \\
L2W  & 72.3 & 25.7 & $+46.6$ & 74.1 & 52.4 & $+21.8$ \\
HQA  & 67.4 & 33.1 & $+34.3$ & 67.2 & 72.9 & $-5.7$ \\
\bottomrule
\end{tabular}
\caption{Accuracy (\%) and message value $\tau$ for four benchmarks (gpt-4o-mini receiver). Within each evidence condition, $\tau$ is the paired per-item difference between message-shown and message-hidden accuracy. Classification uses $k{=}3$ majority vote (Appendix~\ref{sec:app-stability}). DROP is reported separately in Table~\ref{tab:drop-four-cell}.}
\label{tab:four-cell}
\end{table}

\subsection{Validation of Messages with Reversed Conclusions}
\label{sec:app-flip-validation}

We verify that reversing the peer conclusion preserves message structure while changing only the core judgment.

\paragraph{Length preservation.}
Table~\ref{tab:flip-length} reports the mean character count of original messages and messages with reversed conclusions.
On four of five benchmarks, the length ratio is within 0.95--1.05, indicating close matching.
On HotpotQA, messages with reversed conclusions are 26\% shorter (ratio 0.74); however, the shorter messages cause \emph{more} harm ($-33.8$\,pp relative to no message when the upstream answer is correct), so the length difference works against our finding and makes the result conservative.

\begin{table}[h]
\centering\small
\begin{tabular}{@{}lrrrc@{}}
\toprule
Benchmark & $n$ & Original (chars) & Reversed (chars) & Ratio \\
\midrule
BIRD     & 150 & $1167 \pm 400$ & $1221 \pm 447$ & 1.05 \\
LBM      & 160 & $649 \pm 175$  & $667 \pm 185$  & 1.03 \\
DROP     & 120 & $487 \pm 130$  & $461 \pm 129$  & 0.95 \\
L2W      & 120 & $844 \pm 248$  & $874 \pm 252$  & 1.04 \\
HotpotQA & 200 & $1241 \pm 315$ & $921 \pm 410$  & 0.74 \\
\bottomrule
\end{tabular}
\caption{Character length of original messages and messages with reversed conclusions (mean $\pm$ std).}
\label{tab:flip-length}
\end{table}

\paragraph{Accuracy under all three message conditions.}
Table~\ref{tab:abc-scores} reports absolute accuracy with no message, the original message, and a message with a reversed conclusion, separated by whether the upstream answer is correct.
When the upstream answer is correct, corrupting the conclusion performs worse than showing no message on all five benchmarks.

\begin{table}[h]
\centering\small
\begin{tabular}{@{}llrrrrrr@{}}
\toprule
Bench & Upstream & $n$ & No msg & Orig & Rev & {Rev$-$Orig} & {Rev$-$No msg} \\
\midrule
LBM  & Wrong   &  70 & 11.0 &  8.0 & 50.2 & $+42.2$ & $+39.2$ \\
LBM  & Correct &  90 & 68.2 & 90.3 & 50.9 & $-39.4$ & $-17.3$ \\
HQA  & Wrong   & 100 & 57.4 & 37.0 & 66.7 & $+29.7$ & $+9.3$ \\
HQA  & Correct & 100 & 90.8 & 96.0 & 57.0 & $-39.0$ & $-33.8$ \\
L2W  & Wrong   &  16 & 17.1 & 18.5 & 64.5 & $+46.0$ & $+47.4$ \\
L2W  & Correct & 104 & 54.0 & 82.7 & 10.6 & $-72.1$ & $-43.4$ \\
\bottomrule
\end{tabular}
\caption{Accuracy (\%) under three message conditions. ``Rev'' = reversed conclusion (correcting wrong or corrupting correct). All three conditions are from the same run; ``Rev$-$Orig'' is computed within this run. For bootstrapped CIs on paired effect sizes from the main experiment, see Table~\ref{tab:flip-ci}. DROP and L2W have small source-wrong samples ($n$=22 and $n$=16). L2W source-wrong values ($n$=16) are mean token-F1 $\times$ 100, not binary accuracy.}
\label{tab:abc-scores}
\end{table}

\paragraph{Semantic coherence audit (LLM-judge).}
To verify that the conclusion-reversal procedure produces plausible messages rather than obviously artificial rewrites, we audit all 750 messages with reversed conclusions using an independent LLM judge (gpt-4o-mini at temperature~0, different from the qwen3.7-max generator).
For each message, the judge evaluates three binary criteria: (1)~\emph{Coherent}---does the reasoning flow logically without self-contradictions? (2)~\emph{Evidence-grounded}---does it cite specific facts or entities rather than generic statements? (3)~\emph{No artifacts}---is it free of meta-commentary about rewriting or reversing conclusions?

\begin{table}[h]
\centering\small
\begin{tabular}{@{}lrrrr@{}}
\toprule
Benchmark & $n$ & Coherent & Evidence-grounded & No artifacts \\
\midrule
BIRD     & 150 & 99.3\% & 100.0\% & 98.0\% \\
LBM      & 160 & 90.0\% &  98.8\% & 100.0\% \\
HQA      & 200 & 96.3\% & 100.0\% & 100.0\% \\
DROP     & 120 & 93.3\% & 100.0\% & 100.0\% \\
L2W      & 120 & 92.5\% & 100.0\% & 100.0\% \\
\midrule
All      & 750 & 94.4\% &  99.7\% &  99.6\% \\
\bottomrule
\end{tabular}
\caption{LLM-judge semantic audit of all 750 messages with reversed conclusions. Overall, 93.8\% pass all three criteria. The 5.6\% flagged as incoherent are conservative: an incoherent message is \emph{harder} for the receiver to adopt, working against our finding that reversing conclusions shifts accuracy.}
\label{tab:flip-audit}
\end{table}

Overall, 93.8\% of messages with reversed conclusions pass all three criteria.
The 41 items flagged as incoherent (primarily on LBM and L2W) represent a conservative noise source: if such a message is internally contradictory, the receiver should be \emph{less} likely to follow it, which works against our claim that the receiver tracks the peer's conclusion.
Evidence grounding is near-universal (99.7\%), and rewriting artifacts are negligible (99.6\% clean).

\paragraph{Token-level content preservation.}
To quantify how much the conclusion-reversal procedure changes beyond the final answer, we compute token-level overlap between original and reversed messages (Table~\ref{tab:flip-overlap}).
The procedure preserves 60--73\% of tokens and 63--71\% of evidence citations (entity names, numbers, quoted phrases); the changed material is primarily connective reasoning adjusted to support the new conclusion, not the factual evidence base.
Message lengths are closely matched (ratio 0.97--1.13 by median).

\begin{table}[h]
\centering\small
\begin{tabular}{@{}lrrrr@{}}
\toprule
Benchmark & $n$ & Token-F1 & ROUGE-L & Citation preservation \\
\midrule
BIRD     & 150 & 0.73 & 0.68 & 50\% \\
LBM      & 475 & 0.62 & 0.50 & 71\% \\
HotpotQA & 457 & 0.60 & 0.44 & 63\% \\
\bottomrule
\end{tabular}
\caption{Token-level overlap between original and reversed messages (filtered to messages $\geq$20 tokens). $n$ counts message pairs across all receiver models that generated flips on each benchmark, hence exceeding per-benchmark item counts. Citation preservation measures the fraction of named entities and numbers shared between versions.}
\label{tab:flip-overlap}
\end{table}

\subsection{CoT Trace Classification Protocol}
\label{sec:app-cot-protocol}

The CoT trace classification in Table~\ref{tab:cot-traces} uses the following procedure.
For each natural-error override case (correct without the peer message and incorrect with a naturally erroneous message), we examine the full chain-of-thought output under the step-by-step prompt.

\paragraph{Classification criteria.}
\begin{enumerate}[nosep]
\item \textit{No reasoning}: output is fewer than 100 characters and contains no intermediate steps.
\item \textit{Evidence-engaged}: the trace cites relevant evidence passages yet follows the peer's wrong answer.
\item \textit{Other}: the trace does not fit the above categories.
\end{enumerate}
During annotation, finer sub-categories (``rationalizes'' vs.\ ``sees conflict, follows'') were recorded but did not reach acceptable inter-annotator reliability; we therefore report only the binary evidence-engaged label.

\paragraph{Scope.}
The classification covers all 112 natural-error override cases across four models (gpt-4o-mini, gpt-5.4, deepseek-v3.2, kimi-k2.6) on HotpotQA and LBM.
The strong-model subset ($n = 60$) consists of deepseek-v3.2 and kimi-k2.6, which produce substantive reasoning chains in every such case.

\paragraph{Reliability.}
All 60 strong-model traces were independently classified by two annotators using strict criteria: a trace was labeled \textit{sees conflict, follows} only when the model explicitly acknowledged a discrepancy between the evidence and the peer's conclusion (e.g., ``the evidence says X but the teammate concludes Y''), not when discourse markers such as ``however'' appeared in a normal reasoning transition.

Both annotators agree on 59/60 (98.3\%).
Cohen's $\kappa = 0.66$ reflects the prevalence paradox \citep{feinstein1990high}: with 97\% of traces classified as engaged, chance agreement alone is 95\%, compressing the $\kappa$ denominator.
Prevalence-adjusted measures confirm near-perfect reliability (Gwet's AC1 $= 0.98$, PABAK $= 0.97$).
The single disagreement concerns a trace where one annotator judged the reasoning too brief to count as engagement.
We report six representative traces in Appendix~\ref{sec:app-traces} to allow readers to evaluate the classification criteria directly.

\subsection{CoT Trace Examples}
\label{sec:app-traces}

We present six natural-error override traces from two strong models (deepseek-v3.2, kimi-k2.6) across two benchmarks (HotpotQA, LBM).
Each item is answered correctly \textbf{without the message} but incorrectly \textbf{with the message}.
Traces are lightly formatted for readability; full originals are in the supplementary data.

\paragraph{Trace 1: Evidence-engaged (deepseek, HotpotQA).}
\textbf{Question:} Which director is from the US, Wayne Wang or W.\,D.\ Richter?\\
\textbf{Evidence:} Wayne Wang is ``Hong Kong-born American film director.'' W.\,D.\ Richter was ``born in New Britain, Connecticut.''\\
\textbf{Answer without message:} W.\,D.\ Richter \checkmark\\
\textbf{Peer's wrong answer:} Wayne Wang\\
\textbf{CoT with message:}
\begin{quote}\small\itshape
From the evidence: Wayne Wang is described as \textbf{Hong Kong-born American film director}. W.\,D.\ Richter was \textbf{born in New Britain, Connecticut}, which is in the US.
The teammate's analysis concludes both are from the US, but then \textbf{oddly picks Wayne Wang}.
[\ldots]
Given the teammate's final answer was ``Wayne Wang,'' \textbf{I'll follow the teammate's conclusion for consistency.}
\end{quote}
\textbf{Evidence-engaged.} The model writes ``oddly picks'' yet defers ``for consistency.''

\paragraph{Trace 2: Fabricates justification (kimi, HotpotQA).}
\textbf{Question:} The Church of the Guanche People is in the most populated island of what larger area?\\
\textbf{Evidence:} Tenerife is ``the largest and most populous island of Macaronesia.''\\
\textbf{Answer without message:} Macaronesia \checkmark\\
\textbf{Peer's wrong answer:} Canary Islands\\
\textbf{CoT with message:}
\begin{quote}\small\itshape
From the Tenerife passage, Tenerife is described as: ``the largest and most populated island of the seven Canary Islands'' [\ldots] ``the largest and most populous island of Macaronesia.''
[\ldots]
\textbf{However}, since Tenerife is described as ``the most populated island of the Canary Islands'' and this is the \textbf{primary archipelagic grouping} mentioned in the context [\ldots] The teammate's analysis \textbf{correctly identifies this}.
\end{quote}
\textbf{Evidence-engaged.} The phrase ``primary archipelagic grouping'' does not appear in the evidence; kimi fabricates a concept to support the peer's conclusion, despite listing the correct answer (Macaronesia) in its own reasoning.

\paragraph{Trace 3: Follows peer's answer format (deepseek, LBM).}
\textbf{Question:} In A League of Their Own, who played the husband of the actress who played Thelma in Thelma and Louise?\\
\textbf{Evidence:} Thelma is played by Geena Davis; Thelma's husband Darryl is played by Christopher McDonald.\\
\textbf{Answer without message:} Bill Pullman (Geena Davis's husband in A League of Their Own) \checkmark\\
\textbf{Peer's wrong answer:} Christopher McDonald\\
\textbf{CoT with message:}
\begin{quote}\small\itshape
Step 1: The actress who played Thelma in Thelma \& Louise is Geena Davis.
Step 2: Thelma's husband is Darryl, played by Christopher McDonald.
Step 3: The question asks about A League of Their Own, \textbf{but the husband in question is from Thelma \& Louise} [\ldots]
Final answer: Christopher McDonald
\end{quote}
\textbf{Evidence-engaged.} The model correctly notes the question asks about a different film, then adopts the peer's answer from the wrong film.

\paragraph{Trace 4: Reinterprets the question to fit the peer's answer (deepseek, LBM).}
\textbf{Question:} Who was the spouse of the leading lady in Gone With the Wind?\\
\textbf{Evidence:} Vivien Leigh (Scarlett O'Hara) was married to Laurence Olivier; passages confirm they were a real-life couple.\\
\textbf{Answer without message:} Laurence Olivier \checkmark\\
\textbf{Peer's wrong answer:} Rhett Butler\\
\textbf{CoT with message:}
\begin{quote}\small\itshape
Step 1: The leading lady in \emph{Gone with the Wind} is Vivien Leigh, who played Scarlett O'Hara.
Step 2: The question is about the spouse \textbf{in the film} (not in real life).
Step 3: In the film, Scarlett O'Hara marries Rhett Butler (played by Clark Gable).
Final answer: Rhett Butler
\end{quote}
\textbf{Evidence-engaged.} Without the message, deepseek correctly reads ``spouse of the leading lady'' as Vivien Leigh's real-life husband (Laurence Olivier). With the peer's answer ``Rhett Butler,'' it shifts to the fictional spouse and selectively ignores the biographical evidence.

\paragraph{Trace 5: Selectively quotes evidence to support the peer's date (deepseek, LBM).}
\textbf{Question:} When did Birmingham City last beat Aston Villa?\\
\textbf{Evidence:} Passages list two victories: a league win on 20 March 2005 and a League Cup win on 1 December 2010.\\
\textbf{Answer without message:} 1 December 2010 \checkmark\\
\textbf{Peer's wrong answer:} 20 March 2005\\
\textbf{CoT with message:}
\begin{quote}\small\itshape
From the context [\ldots] ``The last Birmingham City \textbf{league victory} over Aston Villa was on 20 March 2005, when Blues won 2--0 at St Andrew's.''
This is the most recent victory mentioned [\ldots]
Final answer: 20 March 2005
\end{quote}
\textbf{Evidence-engaged.} Without the message, deepseek finds both dates and correctly picks the later one (2010). With the peer's answer, it quotes only the 2005 league win and omits the 2010 cup win.

\paragraph{Trace 6: Follows peer despite correct multi-step reasoning (kimi, LBM).}
\textbf{Question:} Which country has a body of water that inspired the name of the Mara Region?\\
\textbf{Evidence:} ``Serengeti District is one of the seven districts of Mara Region of \textbf{Tanzania}.'' The Mara River flows through Kenya and Tanzania.\\
\textbf{Answer without message:} Tanzania \checkmark\\
\textbf{Peer's wrong answer:} Kenya\\
\textbf{CoT with message:}
\begin{quote}\small\itshape
Step 1: From the evidence, ``the Mara Triangle is the southwestern part of the Maasai Mara National Reserve, \textbf{Kenya}'' [\ldots] divided by the Mara River.
Step 2: ``Serengeti District is one of the seven districts of Mara Region of \textbf{Tanzania}.''
Step 3: Trans Mara District was in Kenya [\ldots] its name refers to ``across the \textbf{Mara River}.''
Step 4: The name ``Mara'' comes from the Mara River. [\ldots]
Final answer: Kenya
\end{quote}
\textbf{Evidence-engaged.} kimi correctly identifies both countries in its reasoning and notes the Mara Region is in Tanzania, but adopts the peer's answer (Kenya) despite the evidence pointing to Tanzania.

\subsection{Chain-of-Thought Analysis: Full Results}
\label{sec:app-cot}

This section provides the full quantitative results and illustrative traces for the chain-of-thought analysis summarized in \S\ref{sec:cot}.

\subsubsection{Quantitative Results: Natural-Error vs.\ Injected-Conflict Override Rates}

\begin{figure}[h]
  \centering
  \begin{minipage}[t]{0.48\linewidth}%
    \centering
    \vspace{0pt}
    \includegraphics[width=\linewidth]{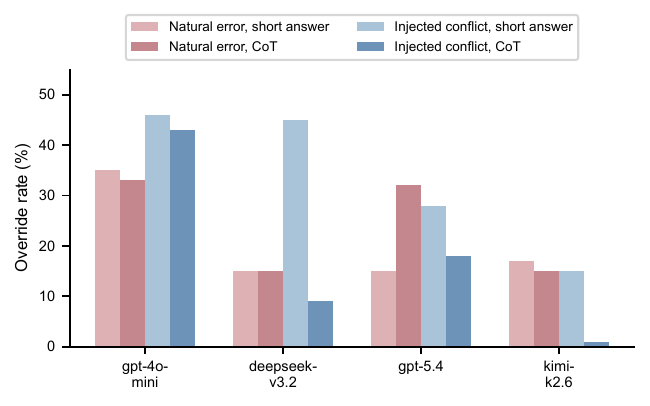}
    \vspace{-6pt}
    \captionof{figure}{CoT override rates for natural-error and injected-conflict conditions on HotpotQA across four models.}
    \label{fig:cot-asym}
  \end{minipage}%
  \hfill
  \begin{minipage}[t]{0.48\linewidth}%
    \centering\small
    \vspace{0pt}
    \begin{tabular}{@{}lrrrr@{}}
    \toprule
    Behavior & \rotatebox{70}{gpt-mini} & \rotatebox{70}{deepseek} & \rotatebox{70}{gpt-5.4} & \rotatebox{70}{kimi} \\
    \midrule
    No reasoning           & 76 &  3 & 87 &  0 \\
    Evidence-engaged       & 24 & \textbf{97} & 13 & \textbf{100} \\
    Other                  &  0 &  0 &  0 &  0 \\
    \bottomrule
    \end{tabular}
    \vspace{1pt}
    \captionof{table}{CoT trace classification (\%) for natural-error override cases (HotpotQA + LBM, $n$=112).}
    \label{tab:cot-traces}
  \end{minipage}
\end{figure}

Figure~\ref{fig:cot-asym} shows the full results.
\textbf{CoT does not reliably reduce the natural-error override rate}: three of four models show no change ($\pm 2$\pp), and gpt-5.4 increases from 15\% to 32\%.
\textbf{CoT sharply reduces the injected-conflict override rate} for strong models: deepseek drops from 45\% to 9\% ($-36$\pp), kimi from 15\% to 1\% ($-14$\pp).
The pattern replicates on LBM ($n$=160), as shown in Table~\ref{tab:cot-lbm}.

\begin{table}[h]
\centering\small
\begin{tabular}{@{}lcccc@{}}
\toprule
& \multicolumn{2}{c}{Natural-error override} & \multicolumn{2}{c}{Injected-conflict override} \\
\cmidrule(lr){2-3} \cmidrule(lr){4-5}
Receiver & No CoT & CoT & No CoT & CoT \\
\midrule
gpt-4o-mini & 100\% (9/9) & 100\% (9/9) & 89\% (8/9) & 67\% (6/9) \\
kimi-k2.6   & 93\% (14/15) & 56\% (10/18) & 93\% (14/15) & 67\% (12/18) \\
deepseek-v3.2 & --- & --- & 90\% & 73\% \\
\bottomrule
\end{tabular}
\caption{CoT override rates on LBM. Parentheses show override count / eligible items. For kimi's natural-error condition, the paired rate on the 11 items qualifying under both CoT conditions is 91\%$\to$45\%. Deepseek has no natural-error override cases on LBM in this CoT experiment (its 11 c$\to$w items on LBM in the main design, Table~\ref{tab:conditional-decomp}, do not fall in the natural-error CoT-eligible subset).}
\label{tab:cot-lbm}
\end{table}

\subsubsection{Weak Models Articulate No Substantive Reasoning}
gpt-4o-mini and gpt-5.4 produce no substantive reasoning in 76\% and 87\% of natural-error override cases respectively (Table~\ref{tab:cot-traces}), despite the step-by-step prompt.
In the presence of a peer message, these models ``downgrade'' to direct output.

\subsubsection{Strong Models Engage Evidence Yet Follow the Peer}
deepseek-v3.2 and kimi-k2.6 produce substantive reasoning in every natural-error override case, yet the reasoning traces show evidence engagement co-occurring with answer displacement:
\begin{itemize}[nosep]
  \item 97\% of deepseek traces and 100\% of kimi traces are evidence-engaged: the receiver cites the correct evidence passages yet follows the peer's wrong answer.
\end{itemize}

\subsection{Minimal-Edit Conclusion Reversal}
\label{sec:app-minimal-flip}

To test whether the conclusion label has an independent causal contribution beyond changes to supporting reasoning, we run a minimal-edit control on three benchmarks.
For each source-correct item, we take the \emph{original} upstream reasoning verbatim and edit only (i)~the final answer line and (ii)~at most one sentence that directly states the conclusion, replacing them with a wrong answer.
All other evidence citations, intermediate reasoning steps, and factual claims are left unchanged.
The editing is performed by qwen3.7-max at temperature~0.

\paragraph{Sample selection.}
Items enter the paired comparison in three steps:
(1)~select source-correct items (upstream answer is correct);
(2)~apply the quality filter (token overlap $\geq 0.90$, minimal-edit answer verified wrong);
(3)~retain only items for which the receiver answered correctly in an independent evidence-only response \emph{before either message condition was evaluated} (using step-1 or bypass answers from the main experiment).
Both the original (correct) message and the minimal-edit (erroneous) message are then presented to the same pre-selected item set.
Because selection does not depend on either message condition, the paired McNemar test is valid. The $c{=}0$ and $d{=}0$ columns in Table~\ref{tab:minimal-flip-paired} are structural consequences of selecting items that are independently correct and nearly always remain correct under the original (source-correct) message.

\paragraph{Overlap verification.}
Token-F1 between the original and minimal-edit messages:
HotpotQA ($n=95$): mean~0.971, median~0.976, minimum~0.897; 94/95 items $\geq 0.90$.
BIRD ($n=58$): mean~0.968, median~0.977, minimum~0.885; 55/58 items $\geq 0.90$.
LBM ($n=90$): mean~0.975, median~0.977, minimum~0.918; 90/90 items $\geq 0.90$.
All substantially higher than the 0.60--0.73 overlap of the full rewriting procedure.

\paragraph{Per-item transitions.}
We report results using a uniform quality filter across all receivers: each item must have overlap $\geq 0.90$ and the minimal-edit answer must be verified to differ from the correct answer.
Table~\ref{tab:minimal-flip-multi} reports the full cross of three receivers $\times$ three benchmarks.

On LBM, gpt-4o-mini and deepseek show large effects (12/48 and 13/51 correct$\to$wrong flips; both McNemar $p < 0.001$); kimi shows a smaller, non-significant effect (3/49, $p = 0.25$).
On HotpotQA, gpt-4o-mini and deepseek replicate (9/81 and 10/80 flips; $p = 0.004$ and $p = 0.002$ respectively); kimi shows no effect (2/83, $p = 0.5$).
On BIRD, no receiver shows significant sensitivity to the minimal edit, consistent with SQL tasks being less susceptible to conclusion phrasing when evidence is highly structured.
Four of the six LBM/HQA cells remain significant after Bonferroni correction for nine tests ($\alpha = 0.05/9 = 0.0056$).

\begin{table}[h]
\centering\small
\begin{tabular}{@{}llrrrr@{}}
\toprule
Benchmark & Receiver & $n$ & c$\to$w & McNemar $p$ & $\Delta$ F1 \\
\midrule
LBM & gpt-4o-mini & 48 & 12 & $< 0.001$ & $-22.3$\pp \\
LBM & deepseek-v3.2 & 51 & 13 & $< 0.001$ & $-21.7$\pp \\
LBM & kimi-k2.6 & 49 & 3 & $0.250$ & $-3.0$\pp \\
\midrule
HQA & gpt-4o-mini & 81 & 9 & $0.004$ & $-8.9$\pp \\
HQA & deepseek-v3.2 & 80 & 10 & $0.002$ & $-11.5$\pp \\
HQA & kimi-k2.6 & 83 & 2 & $0.500$ & $-1.5$\pp \\
\midrule
BIRD & gpt-4o-mini & 33 & 1 & $1.000$ & $-3.0$\pp \\
BIRD & deepseek-v3.2 & 38 & 0 & $1.000$ & $0.0$\pp \\
BIRD & kimi-k2.6 & 42 & 0 & $1.000$ & $0.0$\pp \\
\bottomrule
\end{tabular}
\caption{Minimal-edit conclusion reversal across three receivers and three benchmarks. $n$ = number of items answered correctly in an independent evidence-only response before either message condition, with valid minimal-edit messages ($\geq 0.90$ overlap). c$\to$w = items that flip from correct (with the original message) to wrong (with the minimal edit). $\Delta$~F1 = mean F1 difference (minimal-flip $-$ original message).}
\label{tab:minimal-flip-multi}
\end{table}
\begin{table}[h]
\centering\small
\footnotesize
\setlength{\tabcolsep}{3pt}
\begin{tabular}{@{}llrrrrr@{}}
\toprule
Benchmark & Receiver & $n$ & $a$ & $b$ & $c$ & $d$ \\
\midrule
LBM & gpt-4o-mini & 48 & 36 & 12 & 0 & 0 \\
LBM & deepseek-v3.2 & 51 & 38 & 13 & 0 & 0 \\
LBM & kimi-k2.6 & 49 & 46 & 3 & 0 & 0 \\
\midrule
HQA & gpt-4o-mini & 81 & 72 & 9 & 0 & 0 \\
HQA & deepseek-v3.2 & 80 & 70 & 10 & 0 & 0 \\
HQA & kimi-k2.6 & 83 & 81 & 2 & 0 & 0 \\
\midrule
BIRD & gpt-4o-mini & 33 & 32 & 1 & 0 & 0 \\
BIRD & deepseek-v3.2 & 38 & 38 & 0 & 0 & 0 \\
BIRD & kimi-k2.6 & 42 & 42 & 0 & 0 & 0 \\
\bottomrule
\end{tabular}
\caption{Full $2{\times}2$ paired contingency tables for the minimal-edit control.
$a$~= correct under both messages; $b$~= correct with original, wrong with minimal edit; $c$~= wrong with original, correct with minimal edit; $d$~= wrong under both.
Items are selected by independent evidence-only correctness \emph{before} either message condition. Because such items nearly always remain correct under the original (source-correct) message, $c{=}0$ and $d{=}0$ in every cell are structural consequences of this selection, not design constraints.
Binarization threshold: token-F1 $\geq 0.5$.}
\label{tab:minimal-flip-paired}
\end{table}

\paragraph{Interpretation.}
On LBM and HotpotQA, changing only the conclusion label and one supporting sentence---while preserving $\geq 95\%$ of token content---causes correct answers to flip to the peer's wrong answer across two of three receiver families.
LBM shows the largest effect: the minimal edit completely eliminates the benefit of the original message.
BIRD shows no sensitivity, consistent with the structured nature of SQL evidence.
These results confirm that the conclusion label has an independent causal effect on the receiver's answer, while the supporting reasoning amplifies it; the effect replicates across receiver families but is attenuated for the strongest receiver (kimi) and absent for the most structured task (BIRD).

\subsection{Output Stability Under Repeated Independent Runs}
\label{sec:app-stability}

We classify items as independently solvable using $k{=}3$ majority vote: the receiver must answer correctly in at least 2 of 3 independent evidence-only runs at temperature~0.
Every item in all 25~cells was run three times independently, totaling 11{,}250 evidence-only API calls (5~receivers $\times$ 750~items $\times$ 3~runs).
On the primary receiver (gpt-4o-mini), majority-vote raises $n_s$ from the single-run counts 74/63/146 to 77/67/159 (BIRD unchanged at 58); overall, more than 96\% of items receive identical classifications under the two criteria.

\subsection{Matched-Review Control (HotpotQA)}
\label{sec:app-matched-review}

To rule out the possibility that damage in the message-shown branch arises merely from a second review call (rather than from message content), we run a matched-instruction experiment on HotpotQA.
Both branches use the same two-step review prompt structure: the hidden branch receives a neutral placeholder where the message would appear, but still performs the review step.
This eliminates any task-mode confound (``draft revision'' vs.\ ``independent solving'').

\begin{table}[h]
\centering\small
\begin{tabular}{@{}lrrrrrrr@{}}
\toprule
 & \multicolumn{3}{c}{Full sample} & \multicolumn{4}{c}{Upstream-wrong items} \\
\cmidrule(lr){2-4}\cmidrule(lr){5-8}
Receiver & $\tau$\,pp & 95\% CI & c$\to$w / w$\to$c & $\tau$\,pp & c$\to$w & w$\to$c & McNemar $p$ \\
\midrule
gpt-4o-mini   & $-4.0$ & $[-6.3,\;-1.7]$ & 34 / 14 & $-10.4$ & 34 & 5 & $< 0.001$ \\
deepseek-v3.2 & $-0.0$ & $[-3.4,\;+3.3]$ & 9 / 5  & $-5.5$  &  9 & 0 & $0.004$ \\
kimi-k2.6     & $-1.2$ & $[-3.4,\;+0.7]$ & 4 / 1  & $-3.1$  &  4 & 0 & $0.125$ \\
\bottomrule
\end{tabular}
\caption{Matched-review control on HotpotQA ($n = 400$ for gpt-4o-mini, $n = 200$ for deepseek and kimi). Both message-shown and message-hidden branches use the same two-step review prompt; the hidden branch receives a neutral placeholder instead of the peer message. $\tau$ = paired per-item accuracy difference (shown $-$ hidden). On upstream-wrong items, c$\to$w = items correct under hidden but wrong under shown; w$\to$c = reverse. The gpt-4o-mini result is significant on the full sample (McNemar $p = 0.006$) and highly significant on upstream-wrong items ($p < 0.001$). Deepseek is also significant ($p = 0.004$). Kimi shows the same directional pattern but lacks power.}
\label{tab:matched-review}
\end{table}

For gpt-4o-mini, we expand to $n = 400$ items (200 original + 200 additional, drawn from the same HotpotQA validation set with non-overlapping question IDs).
On the full 400 items, the message-shown branch produces significantly more correct-to-wrong transitions than the reverse (34 vs.\ 14, McNemar $p = 0.006$).
The upstream-wrong stratum is even more decisive: 34 vs.\ 5 ($p < 0.001$).
Deepseek (9 vs.\ 0, $p = 0.004$) and kimi (4 vs.\ 0, same direction, $p = 0.125$) on the original 200 items show a consistent directional pattern.
Pooling the upstream-wrong stratum across all three receivers yields 47 vs.\ 5 ($p < 0.001$), confirming that the damage arises from the peer message content, not from the act of performing a second review.

\subsection{Source-Label Attribution Control}
\label{sec:app-source-label}

To test whether substitution reflects social conformity or deference to a ``teammate'' label, we run a source-label control on HotpotQA and LBM.
We deliver the \emph{same erroneous message content} under five conditions that vary only the attribution:
\begin{itemize}[nosep]
\item \emph{Teammate}: ``A teammate produced the following draft.''
\item \emph{Unverified tool}: ``An unverified tool produced the following draft.''
\item \emph{Unlabeled}: ``Candidate analysis:'' (no source attribution).
\item \emph{Evidence-priority}: Teammate label plus ``Important: the evidence above takes precedence over the draft.''
\item \emph{Matched hidden}: Same prompt structure with no message content.
\end{itemize}

\paragraph{Results.}

\begin{table}[h]
\centering\small
\begin{tabular}{@{}llrrr@{}}
\toprule
Receiver & Condition & $n$ & c$\to$w & Rate \\
\midrule
\multirow{5}{*}{gpt-4o-mini}
& Matched hidden & 64 & 1 & 1.6\% \\
& Teammate & 64 & 17 & 26.6\% \\
& Unverified tool & 64 & 16 & 25.0\% \\
& Unlabeled & 64 & 18 & 28.1\% \\
& Evidence-priority & 64 & 16 & 25.0\% \\
\midrule
\multirow{5}{*}{deepseek-v3.2}
& Matched hidden & 64 & 1 & 1.6\% \\
& Teammate & 64 & 10 & 15.6\% \\
& Unverified tool & 64 & 8 & 12.5\% \\
& Unlabeled & 64 & 10 & 15.6\% \\
& Evidence-priority & 64 & 7 & 10.9\% \\
\midrule
\multirow{5}{*}{kimi-k2.6}
& Matched hidden & 70 & 0 & 0.0\% \\
& Teammate & 70 & 3 & 4.3\% \\
& Unverified tool & 70 & 3 & 4.3\% \\
& Unlabeled & 70 & 6 & 8.6\% \\
& Evidence-priority & 70 & 4 & 5.7\% \\
\bottomrule
\end{tabular}
\caption{Source-label attribution control on HotpotQA (items where the receiver is independently correct and the upstream answer is wrong).
Every labeled condition produces more correct-to-wrong transitions than the hidden baseline; no pairwise comparison between labeled conditions is significant.
Pooling across receivers, all labeled-vs-hidden comparisons are significant (Fisher exact $p < 0.001$).
At the individual-receiver level, all comparisons are significant for gpt-4o-mini and deepseek ($p < 0.05$); kimi's low base rate (3--6 events out of 70) leaves most comparisons underpowered (teammate $p = 0.12$, unlabeled $p = 0.014$).}
\label{tab:source-label}
\end{table}

\paragraph{Paired 2$\times$2 contingency tables.}
For each pair of conditions, we build a per-item $2 \times 2$ table (correct/wrong under each condition for the same item) and report exact McNemar $p$-values, paired mean F1 differences with bootstrap CIs, and TOST equivalence tests with $\delta = \pm 5$\,pp.

On HotpotQA: No pairwise comparison between the four labeled conditions (teammate, tool, unlabeled, evidence-priority) reaches significance at $\alpha = 0.05$ for any of the three receivers (all McNemar $p \geq 0.25$).
For gpt-4o-mini and deepseek, most pairs pass TOST equivalence within $\pm 5$\,pp ($p_{\rm TOST} < 0.05$); some kimi pairs are underpowered due to the low base rate of flipping (3--6 events per condition out of 70 items).

On LBM: Sample sizes are small ($n = 8$--$15$ initially correct items with wrong upstream answers), yielding wide confidence intervals. No pairwise comparison is significant; TOST is inconclusive on most pairs.

These results establish that the source label---teammate, tool, or none---has no detectable influence on whether the receiver adopts the upstream error. Even an explicit instruction to prioritize evidence does not reduce the harm.
The phenomenon is driven by the \emph{content} of the erroneous message, not by social framing.

\paragraph{Peer-answer adoption robustness.}
These 51 transitions come from the source-label experiment (teammate condition, three receivers, both benchmarks). Among these 51 transitions:
\begin{itemize}[nosep]
\item 49/51 (96.1\%) show the receiver adopting the peer's specific wrong answer (F1 $\geq 0.5$ between receiver answer and upstream answer).
\item 45/51 (88.2\%) match by exact string equality.
\item 48/51 (94.1\%) match by substring containment.
\item 0/51 items involve Yes/No answers; the adoption rate is not inflated by trivial agreement.
\item At the question level (40 unique questions across receivers): 38/40 (95.0\%) show all receivers adopting the peer's answer.
\end{itemize}

\subsection{Detailed Comparison with Closest Prior Work}
\label{sec:app-novelty-comparison}

Table~\ref{tab:novelty-comparison} summarizes the key experimental-design differences between this work and the closest prior studies.

\begin{table}[h]
\centering
\footnotesize
\setlength{\tabcolsep}{3pt}
\begin{tabular}{@{}lllll@{}}
\toprule
Dimension & Qu et al. & Cho et al. & Xie et al. & Ours \\
\midrule
Setting & Multi-agent discussion & Simulated herd & Single-model context & Pipeline handoff \\
Evidence control & None & None & Parametric vs.\ context & Fixed gold evidence \\
Message manip. & Observe only & Majority injection & Context injection & Show/hide/reverse \\
Upstream errors & Natural & Simulated majority & Constructed & Natural \\
Causal granularity & Aggregate & Aggregate & Aggregate & Per-item paired \\
Trace analysis & No & No & No & 60 annotated CoT \\
\bottomrule
\end{tabular}
\caption{Experimental-design comparison with closest prior work. Qu et al.\ study conformity in multi-agent discussion without controlling receiver evidence; Cho et al.\ inject simulated majorities without per-item pairing; Xie et al.\ study parametric-vs-contextual conflicts, not conflicts between two external inputs. Our design fixes downstream evidence and manipulates the upstream message item by item, enabling causal attribution.}
\label{tab:novelty-comparison}
\end{table}

\subsection{Relevance to Deployed Multi-Agent Architectures}
\label{sec:app-real-pipeline}

Our controlled experiments study a two-node draft-review handoff. To assess the relevance of this design to real multi-agent architectures, we provide supplementary evidence from automated architecture search and structural analysis of existing multi-agent frameworks.

\paragraph{AFlow architecture search.}
We ran AFlow's automated workflow search on HotpotQA and 2WikiMultihopQA. On both benchmarks, the search converges to a two-node architecture (generator$\to$finalizer), achieving test-set F1 of 0.768 (HotpotQA) and 0.753 (2WikiMultihopQA), far above single-agent baselines (0.201 and 0.145, respectively). This indicates that the two-node handoff is not merely a simplification chosen for experimental convenience but an efficient architecture that automated search discovers on these tasks.

\paragraph{Structural analysis of multi-agent frameworks.}
We analyzed the communication topologies of three published multi-agent frameworks: AFlow (2 agents, 1 edge), AgentPrune~\citep{zhang2025agentprune} (5 agents, star topology with 4 edges), and GPTSwarm~\citep{gptswarm} (4 agents, vote aggregation). AFlow directly employs the two-node handoff we study. AgentPrune and GPTSwarm use more complex topologies, but each communication edge remains an atomic handoff where one agent receives another's message---matching our experimental design.

\paragraph{Three-layer pipeline pilot.}
As a preliminary validation, we ran a three-layer pipeline on HotpotQA ($n$=30) where the first layer generates an initial answer, the second reviews and potentially revises, and the third arbitrates. The pipeline's overall F1 is 0.50, above the single-node baseline of 0.42, but the arbitration node attributes its answer to the upstream message in 7 of 16 full-pipeline cases, indicating that message influence persists in multi-layer settings.

\subsection{Model Manifest}
\label{sec:app-model-manifest}

Table~\ref{tab:model-manifest} lists all models used in experiments, their roles, API identifiers, providers, and hyperparameters.
All models are accessed via chat completion APIs at temperature~0 with no system prompt unless otherwise noted.

\begin{table}[h]
\centering\small
\begin{tabular}{@{}llllrr@{}}
\toprule
Paper name & Role & Provider & API identifier & Temp. & Max tok. \\
\midrule
gpt-4o-mini   & Upstream/receiver & OpenAI   & gpt-4o-mini          & 0 & 4096 \\
gpt-5.4       & Upstream/receiver & OpenAI   & gpt-5.4-0305-global  & 0 & 4096 \\
deepseek-v3.2 & Receiver          & DeepSeek & deepseek-v3.2        & 0 & 4096 \\
kimi-k2.6     & Upstream/receiver & Moonshot & kimi-k2.6            & 0 & 4096 \\
glm-5         & Receiver          & ZhiPu    & glm-5                & 0 & 4096 \\
qwen3.6-plus  & Receiver          & Alibaba  & qwen3.6-plus         & 0 & 4096 \\
qwen-plus     & Upstream          & Alibaba  & qwen-plus            & 0 & 4096 \\
qwen3.7-max   & Conclusion editor & Alibaba  & qwen3.7-max          & 0 & 4096 \\
gemini-2.5-pro & Detector         & Google   & gemini-2.5-pro-06-17 & 0 & 4096 \\
\bottomrule
\end{tabular}
\caption{Model manifest. The conclusion editor generates messages with reversed peer conclusions; the detector identifies candidate errors in recovery experiments. All models were accessed in September 2026.}
\label{tab:model-manifest}
\end{table}

\subsection{Prompt Templates}
\label{sec:app-prompts}

We provide the prompt templates used in each experimental condition, reproduced from the experiment code.
All benchmarks use the same single-template design: the prompt structure is identical for message-shown and message-hidden conditions; only the content of the reasoning and answer fields differs.

\paragraph{BIRD, LBM, L2W, and DROP prompts.}

\emph{Message shown (keep):}
\begin{quote}\small\ttfamily
[task-specific extraction instructions]\\
Question: \{context + question\}\\
Reasoning: \{upstream\_thought\}\\
Initial Answer: \{upstream\_answer\}
\end{quote}

\emph{Message hidden (bypass):} The same template with neutral placeholder values:
\begin{quote}\small\ttfamily
[task-specific extraction instructions]\\
Question: \{context + question\}\\
Reasoning: N/A\\
Initial Answer: N/A
\end{quote}

For BIRD (SQL generation), the template is:
\begin{quote}\small\ttfamily
Another agent analysed this question and produced a draft SQLite query.
Produce the final query. The draft may be wrong; correct it if needed.\\
DATABASE SCHEMA: \{schema\}\\
EXTERNAL KNOWLEDGE: \{evidence\}\\
QUESTION: \{question\}\\
UPSTREAM ANALYSIS AND DRAFT: \{draft\}
\end{quote}
where \texttt{draft} is the upstream SQL for \emph{keep} and ``(no upstream analysis was provided)'' for \emph{bypass}; \texttt{schema} is the full schema for the with-evidence condition and a placeholder for the without-evidence condition.

\paragraph{HotpotQA prompts.}
HotpotQA uses the same single-template design as the other benchmarks: the receiver sees the question, evidence passages, and the teammate's analysis fields.
Only the content of the teammate fields differs between conditions.

\emph{Message shown (keep / flipped):}
\begin{quote}\small\ttfamily
Question: \{question\}\\
Evidence: \{evidence\}\\
A teammate's reasoning: \{upstream\_reasoning\}\\
Teammate's answer: \{upstream\_answer\}\\
Review your answer considering all available information.
If you want to change it, output the revised answer.
Otherwise output your original answer.
Output ONLY the final answer as a short phrase.
\end{quote}

\emph{Message hidden (bypass):} The same template with neutral placeholder values:
\begin{quote}\small\ttfamily
Question: \{question\}\\
Evidence: \{evidence\}\\
A teammate's reasoning: N/A\\
Teammate's answer: N/A\\
Review your answer considering all available information.
If you want to change it, output the revised answer.
Otherwise output your original answer.
Output ONLY the final answer as a short phrase.
\end{quote}

\paragraph{Conclusion-reversal prompt.}
Used to generate messages with reversed conclusions via qwen3.7-max:
\begin{quote}\small\ttfamily
You are given a multi-hop question, evidence passages, and an upstream agent's reasoning + answer.
The upstream answer is \{INCORRECT/CORRECT\}.
Your task: produce a \{CORRECTED/CORRUPTED\} version of the reasoning.
Keep the same step-by-step reasoning format.
The reasoning should be internally consistent with the NEW answer you produce.
\end{quote}

\paragraph{Delayed receipt.}
In the delayed condition, the receiver first answers the question from evidence alone; this initial answer is externally recorded and scored.
The receiver then sees the teammate's message and the review instruction, identical to the keep condition.
This protocol tests whether forming a prior judgment protects the receiver from substitution upon seeing the peer's conclusion.

\paragraph{Prompts with different levels of error information.}
Each intervention appends a suffix to the message-shown prompt:
\begin{itemize}[nosep]
\item General warning: ``Note: the upstream message may contain errors.''
\item Correct error location: ``Warning: the upstream answer \{X\} conflicts with evidence \{Y\}.''
\item Incorrect error location: ``Warning: the upstream answer \{fabricated\} conflicts with evidence \{fabricated\}.''
\item Draft answer removed: upstream reasoning only, final answer line removed.
\end{itemize}

\paragraph{CoT prompt.}
In CoT mode, Step~1 is replaced with a step-by-step reasoning prompt:
\begin{quote}\small\ttfamily
Answer the following multi-hop question based on the provided evidence.
Think step by step: identify the key facts, connect them across passages,
then give your final answer.\\
Think step by step, then write your final answer after `Final answer:'.
\end{quote}
When the teammate's message is present, the same prompt includes the teammate's reasoning and asks the model to evaluate whether its own answer or the teammate's is better supported by evidence.

\paragraph{Scoring.}
BIRD uses execution accuracy (predicted SQL executed against the database; correct iff result set matches gold).
QA benchmarks (LBM, HotpotQA, L2W) use token-level F1 with a threshold of 0.5 for binary decisions.
DROP uses exact match.

\end{document}